\documentclass{article}

\usepackage[preprint]{corl_2026} 

\usepackage[numbers]{natbib}
\usepackage{multicol}

\usepackage{enumitem}
\usepackage{booktabs}  
\usepackage{multirow}  
\usepackage{xcolor}    
\usepackage{geometry}
\usepackage{graphicx} 
\usepackage{amsmath}
\usepackage{subcaption}
\usepackage{caption}
\usepackage{wrapfig}

\usepackage{listings}
\usepackage{tcolorbox}
\definecolor{codegray}{rgb}{0.4,0.4,0.4}      
\definecolor{codeblue}{rgb}{0.11,0.27,0.45}   
\definecolor{codered}{rgb}{0.6,0.0,0.0}       
\definecolor{backcolour}{rgb}{0.96,0.96,0.96} 

\lstdefinestyle{academicpython}{
    language=Python,
    backgroundcolor=\color{backcolour},   
    commentstyle=\color{codegray}\itshape, 
    keywordstyle=\color{codeblue}\bfseries, 
    stringstyle=\color{codered},           
    basicstyle=\ttfamily\footnotesize,     
    breakatwhitespace=false,         
    breaklines=true,                 
    captionpos=b,                    
    keepspaces=true,                 
    numbers=left,                   
    numbersep=5pt,                  
    numberstyle=\tiny\color{gray},   
    showspaces=false,                
    showstringspaces=false,
    showtabs=false,                  
    tabsize=4,                       
    frame=lines,                     
    rulecolor=\color{gray!30},       
}

\title{HATS: A Human–Agent Teleoperation System for Multi-Arm Data Collection}

\author{
  {\bf Zesen Lin}$^1$, {\bf Jian-Jian Jiang}$^1$, {\bf Haoming Cen}$^1$, {\bf Xiao-Ming Wu}$^2$, {\bf Dandan Zhang}$^3$, {\bf Wei-Shi Zheng}$^{1,\dagger}$ \\
  \and
  $^1$School of Computer Science and Engineering, Sun Yat-sen University, China \\
  $^2$Nanyang Technological University, Singapore \\
  $^3$Imperial College London, UK \\
}

\vspace{-0.1cm}
\begin{document}

\maketitle

\makeatletter
\begingroup
\renewcommand\@makefntext[1]{\noindent$^\dagger$#1}
\footnotetext{Corresponding author.}
\endgroup
\makeatother

\thispagestyle{empty}
\pagestyle{empty}
\vspace{-0.2cm}

{ 
\centering
\includegraphics[width=0.9\textwidth]{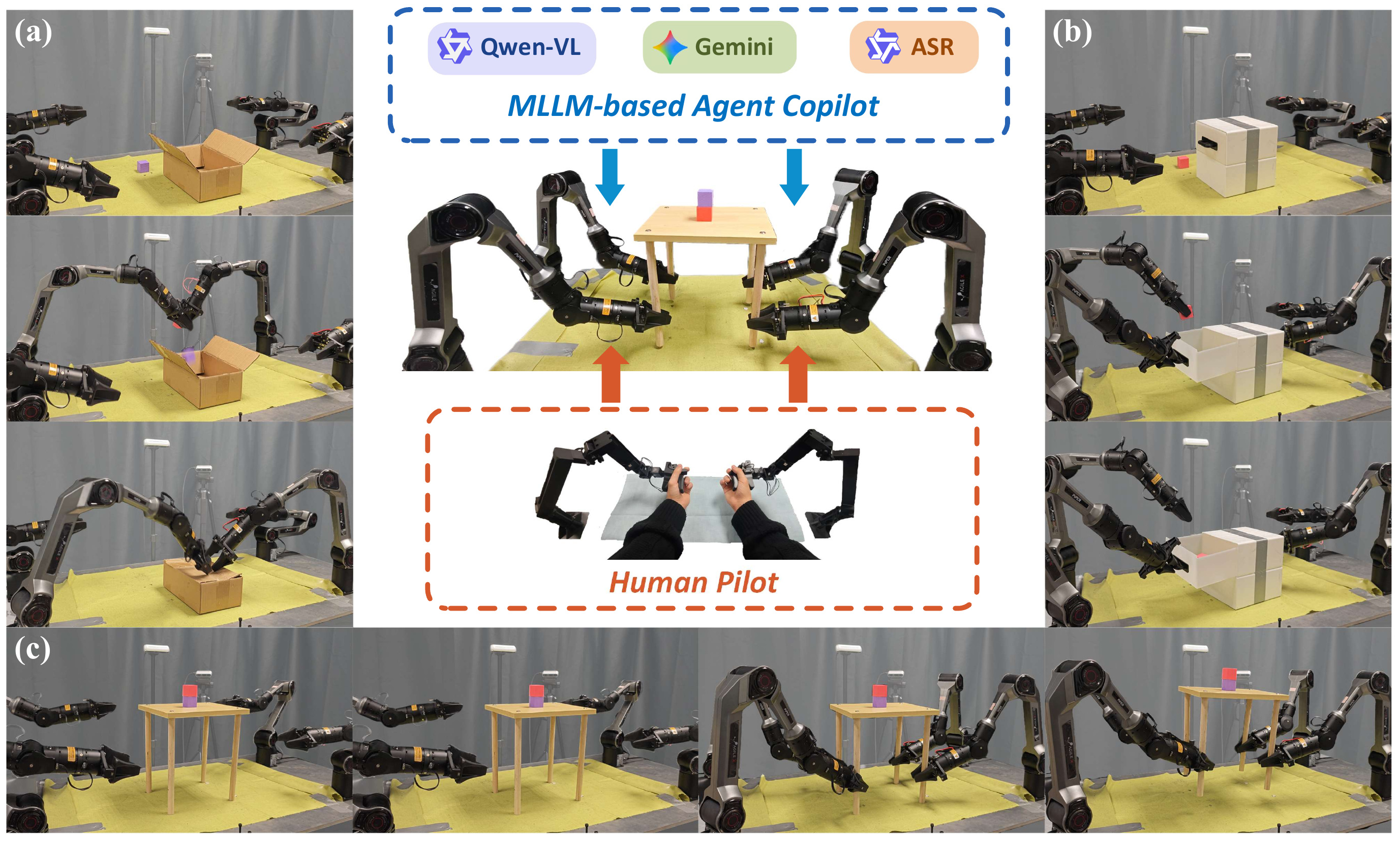}

\vspace{-0.1cm}

\captionof{figure}{\textbf{HATS.} Our human-agent teleoperation system allows a single operator to collect data on four-arm manipulation tasks with an MLLM-based agent. Execution sequences of
three tasks: 
(a): Load items into a box and close the box flaps. 
(b): Open the drawer and put in the object. 
(c): Grasp the four legs of a table and lift it.}
\label{fig:teaser}
\par 
}

\vspace{-0.2cm}


\begin{abstract}
Many real-world manipulation scenarios, such as handling complex collaborative tasks and dealing with large workspaces, require coordination of more than two robotic arms. Consequently, an effective multi-arm teleoperation system is required to collect demonstrations for training coordinated multi-arm manipulation policies.
However, existing teleoperation frameworks mainly focus on single-operator or multi-operator setups, facing a practical trade-off between the cognitive load placed on a single operator and the coordination cost incurred by multiple operators.
To address this problem, we introduce \textbf{HATS}, a human–agent teleoperation system that enables a single human operator, assisted by an MLLM-based agent, to collect data for multi-arm manipulation tasks.
Our system decouples the control space: two primary arms are directly teleoperated by the human, while two assistive arms are controlled by a training-free agent that handles sub-tasks. In addition, the human operator can use voice commands to prevent collisions and correct assistive arm behaviors during execution.
Extensive evaluations demonstrate that HATS achieves data collection efficiency and success rates comparable to expert dual-human teams.
Moreover, downstream policy evaluations demonstrate the efficacy and quality of the data collected through HATS.

\end{abstract}



\section{Introduction}

Many real-world manipulation tasks remain out of reach for bimanual systems because they require coordinated action among more than two arms.
For example, tasks in large workspaces, such as lifting a wide table or spreading out a tablecloth, often require three or four arms to operate simultaneously.
Beyond enabling new capabilities, multi-arm systems can also significantly improve execution speed and task efficiency by parallelizing manipulation and reducing idle time.
Therefore, multi-arm manipulation is essential for robots to operate in such complex scenarios.
Achieving this capability relies on scalable multi-arm teleoperation systems to collect real-world demonstrations for data-driven policy learning \cite{jang2022bc, belkhale2023data}.

However, most existing teleoperation systems are limited to single-arm or dual-arm configurations \cite{qin2023anyteleop, wu2024gello, fang2024airexo, chi2024universal, liu2025factr}, only a few support multi-arm setups \cite{tung2021learning, ozdamar2022shared, kennel2021manipulability, garate2021scalable, kennel2022multi}. Current multi-arm teleoperation frameworks can be broadly grouped into two categories: 1) single-operator control of multi-arm systems imposes excessive cognitive burden, requiring mode-switching between arms and producing discontinuous or inconsistent demonstrations \cite{ozdamar2022shared, garate2021scalable, kennel2021manipulability, kennel2022multi}. 2) multi-operator control reduces individual workload but incurs high labor cost and coordination demands, such as communication delays and inconsistent actions, which may negatively affect the quality of the demonstrations \cite{tung2021learning, huang2020tri}. 

To address these problems, we introduce \textbf{HATS} (\textbf{h}uman–\textbf{a}gent \textbf{t}eleoperation \textbf{s}ystem), a multi-arm human-agent teleoperation system via pre-trained multimodal large language models (MLLMs). The core insight of our approach is to decouple the high-dimensional control space according to the functional roles within the task.
Unlike previous paradigms, we adopt a ``Pilot-Copilot'' paradigm: the human operator serves as the pilot, directly teleoperating two primary arms to perform manipulation, while an assistive agent acts as a copilot, controlling two assistive arms to execute supportive sub-tasks, such as picking, placing, and lifting.
This division of labor enables a single operator to control four arms, reducing human labor cost while maintaining coordinated multi-arm execution.

Our pipeline integrates four modules: perception, planning, collaboration, and supervision. Initially, the perception module extracts the spatial configurations of target objects from visual inputs. Based on this, an LLM planning agent decomposes the task and generates execution plans for the assistive arms. During execution, the assistive arms enact these plans via motion primitives, operating in parallel with the human-controlled primary arms. Concurrently, the supervision module enables the operator to intervene via real-time voice commands, ensuring system adaptability and safety.

To validate the efficacy of our system, we perform experiments on seven real-world tasks. The experimental results indicate that our system achieves a data collection success rate and efficiency comparable to expert dual-human teams, attaining an average success rate of 86.4\%. 

Furthermore, we show that the imitation learning policies trained on HATS data match the performance of those trained on dual-human demonstrations.
These results demonstrate that HATS yields high-quality data, serving as an effective system for multi-arm data collection.

\section{Related Works}
\subsection{From Bimanual to Multi-Arm Teleoperation}
Teleoperation is critical for collecting real-world manipulation data \cite{mandlekar2021matters, fu2024mobile}. While low-cost bimanual systems \cite{zhao2023learning, aldaco2024aloha, wu2024gello} and VR-based retargeting \cite{handa2020dexpilot, qin2023anyteleop, chen2025arcap} have democratized dual-arm data collection, scaling these to multi-arm settings poses severe coordination challenges. Existing multi-arm platforms often require multi-operator collaborations \cite{tung2021learning}, which incur high labor costs and communication latency. To enable single-operator control, previous systems rely on classical Shared Autonomy (SA) formulated via real-time optimization \cite{kennel2022multi} and impedance control \cite{laghi2018shared, manschitz2022shared, fu2025tasc}, or depend on mode-switching mechanisms. This mode-switching paradigm is widely adopted in prominent commercial systems like the Da Vinci surgical system \cite{morrell2021history} and other robotic platforms \cite{ozdamar2022shared, garate2021scalable, kennel2021manipulability}, where a single operator must toggle control between different arms. However, these approaches typically impose rigid symmetric mappings or task-specific constraints, lacking the high-level semantic understanding required for unconstrained, messy real-world environments. In contrast, our system allows a single operator to fluidly control a four-arm system without mode switching. 

To achieve more adaptive assistance, recent SA research has shifted toward data-driven human-in-the-loop paradigms. 
Early works introduced gating mechanisms, where humans intervene only when autonomous policy fails \cite{kelly2019hg, mandlekar2020human}. Recent systems enable agents to learn from real-time interventions \cite{wu2025robocopilot, xu2025hacts} or co-adapt with human operators \cite{luo2025human}. VOSA \cite{belsare2025toward} further explores zero-shot intent recognition using vision-language models.
Nevertheless, these methods are often bottlenecked by the cold-start problem of iterative policy training or the closed-world assumption of predefined goal sets. By leveraging the semantic reasoning capabilities of MLLMs, our assistive agent can decompose language-specified tasks into executable assistive behaviors for diverse manipulation scenarios.


\subsection{MLLMs in Robotics}
MLLMs have been extensively applied to robotic planning, specifically for task decomposition and long-horizon reasoning. Frameworks such as VoxPoser \cite{huang2023voxposer} and Code as Policies \cite{liang2022code} bridge natural language instructions with motion primitives or executable code, facilitating the execution of complex multi-step tasks. Furthermore, recent agent-based architectures, such as ManiAgent \cite{yang2025maniagent}, have explored mapping multimodal observations directly to low-level actions.

However, in high-frequency teleoperation, the use of MLLMs is still limited in practice. 
Real-time deployment raises concerns about latency, reliability, and safety in human-in-the-loop control. 
LAMS \cite{tao2025lams} applies LLMs to infer user intent and switch control modes to reduce operator workload, but does not perform direct kinematic control and is limited to single-arm settings. 
We propose a decoupled collaboration framework in which an MLLM-based agent directly controls assistive arms to execute supportive sub-tasks, while the human operator handles the primary manipulation.


\begin{figure*}[t]
    \centering
    \includegraphics[width=\linewidth]{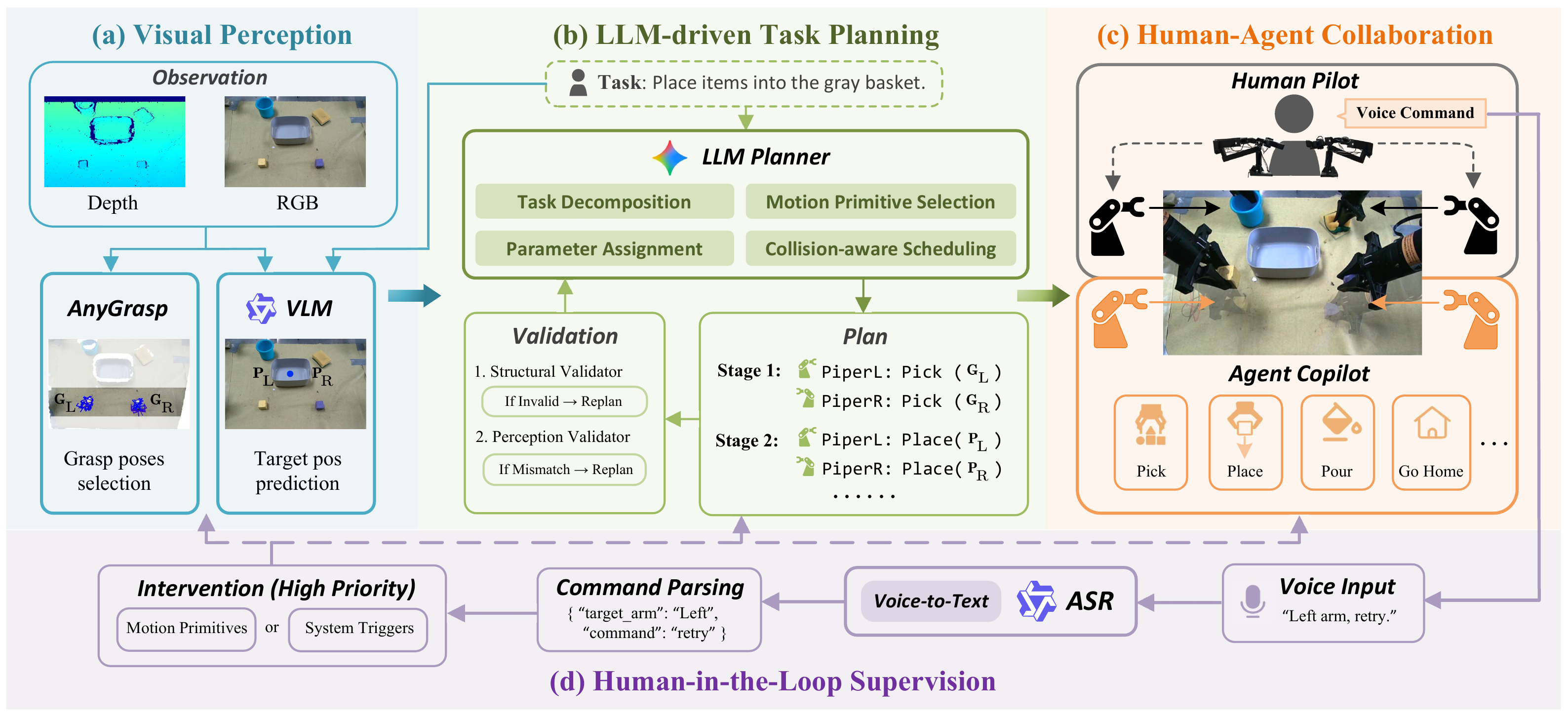}
    \vspace{-0.10cm}
    \caption{\textbf{System Overview.} 
    (a) The perception module extracts grasp poses ($G$) and target positions ($P$) using AnyGrasp and VLM. (b) The planner leverages an LLM planner to generate executable action sequences based on the task description. (c) The collaboration module coordinates the human and agent to control the multi-arm system collaboratively. (d) The supervision module enables the operator to intervene via real-time voice commands, ensuring system adaptability and safety.}
    \label{fig:pipeline}
    \vspace{-0.10cm}
\end{figure*}

\section{Method}
\subsection{System Overview} \label{sec:overview}
As shown in Figure \ref{fig:pipeline}, the system consists of four modules: Perception, Planning, Collaboration, and Supervision. The perception module combines user prompts with visual observations. It uses a 3D camera to capture visual observations, which are then input into the VLM and AnyGrasp \cite{fang2023anygrasp} to extract task-relevant information. 
This information are then passed to the planning module, where the LLM planner first decomposes the task, then assigns subtasks and key parameters to the two assistive arms, selecting the corresponding motion primitives based on the actions to be performed. During collaboration, the operator controls the two primary arms while the two assistive arms automatically perform the task according to the plan. A supervision module with real-time speech recognition allows the operator to intervene or adjust the agent's behavior when necessary.

\subsection{Visual Perception}  \label{sec:perception}
\textbf{Grasp detection.} To ensure robust interaction with diverse and unseen objects, we adopt AnyGrasp, a general-purpose grasp detection algorithm. Given the RGB-D observation, we first convert the depth map into a point cloud $\mathcal{P}_{cam}$ in the camera frame. AnyGrasp processes $\mathcal{P}_{cam}$ to predict a set of candidate grasp poses $\mathcal{G}_{cand} = \{g_1, g_2, \dots, g_N\}$, where each grasp $g_i \in SE(3)$ is parameterized by a translation $T_i \in {R}^3$, a rotation $R_i \in SO(3)$. 
To enhance robustness, we adopt DBSCAN \cite{ester1996density} to cluster the translations $\{T_1, \dots, T_N\}$ of the candidate grasps and identify the dominant cluster $\mathcal{C} \subset \mathcal{G}{cand}$ corresponding to a stable grasping region of the target object.
We then compute the geometric centroid of this cluster to determine the final target position $P_{target} \in {R}^3$:
\begin{equation}
P_{target} = \frac{1}{|\mathcal{C}|} \sum_{g_j \in \mathcal{C}} T_j.
\end{equation}

To determine the target rotation $R_{target}$, we select the candidate $g_k \in \mathcal{C}$ whose translation is closest to the centroid:
\begin{equation}
    k = \operatorname{arg\,min}_{g_j \in \mathcal{C}} \| t_j - P_{\text{target}} \|_2.
\end{equation}
The final grasp pose is therefore defined as $(P_{target}, R_k)$, combining a robust positional estimate with a valid orientation from the nearest neighbor.

\textbf{Semantic placement.} In contrast to grasping,  placement requires grounding high-level linguistic descriptions into the visual scene. To this end, we employ Qwen-VL \cite{bai2023qwen} for open-vocabulary localization of target placement.
Let $\mathcal{L}_{task}$ denote the natural language instruction. We construct a query prompt $\mathcal{Q}$ that combines $\mathcal{L}_{task}$ with the current RGB image $I_{rgb}$. The VLM predicts the pixel coordinates $(u, v)$ corresponding to the centroid of the target placement region. Given the $(u, v)$ and the corresponding depth value $z = D(u, v)$, we recover the 3D point in the camera frame $P_{cam}$ using the camera intrinsic matrix $K$. Finally, the grasp poses $(P_{target}, R_k)$ and the spatial coordinates $P_{cam}$  are forwarded to the planning module as geometric preconditions.

\subsection{LLM-driven Task Planning}  \label{sec:planning}
We employ Gemini 3.1-Flash-lite as the central planner to translate high-level instructions into executable robot actions.
We frame planning as a constrained code generation task, where the LLM planner takes perception results and outputs a structured and readable plan subject to kinematic and temporal constraints.
The planner mainly performs three functions: \textit{task decomposition}, \textit{motion primitive selection}, and \textit{parameter assignment}.
It first understands the task requirements and, based on various constraints, breaks down the original task into staged sub-tasks and assigns specific assistive arms to perform them.
It then selects appropriate motion primitives based on the motion each arm needs to perform and instantiates them with parameters driven by the perception module. 
During both grasping and placement, the planner enforces a collision-aware scheduling rule: when the spatial distance between two targets is below a safety threshold, the corresponding arms have to be executed sequentially.

To ensure robust and adaptive execution, our system adopts a two-stage validation mechanism.
During the initial planning stage, a structural validator checks whether the generated plan satisfies the required schema and parameter constraints. If invalid formatting or parameters are detected, the error feedback is returned to the LLM planner for replanning. Once validated, the plan is cached as a reusable template, requiring only perception-driven parameter updates in subsequent executions, thereby reducing latency and API cost.
Before each data collection episode, a perception-aware validator further verifies whether the current scene matches the cached template by checking perception outputs such as the number of grasp poses. If a mismatch is detected, the system automatically triggers replanning. This design enables efficient repeated data collection while maintaining adaptability to dynamic changes in the scene.

\subsection{Human-Agent Collaboration}  \label{sec:execution}
The function of the collaboration module is to enable human and agent collaboration in completing tasks. On the one hand, it allows the human operator to teleoperate the primary arms using a ``Leader-Follower'' setup; on the other hand, it controls the assistive arms by converting the structured plan into low-level hardware commands.

To support diverse manipulation behaviors, we define a library of motion primitives, which includes not only the original atomic functions of the robotic arm but also the composite behaviors that we define.
At the atomic level, we directly utilize the robot's built-in control functions as the foundation of our system. 
\textbf{Cartesian Control} moves the end-effector to a target pose $(x, y, z, roll, pitch, yaw)$. 
\textbf{Joint Control} sets the six joint angles to a desired configuration. When combined with an inverse kinematics (IK) solver, this allows the arm to reach any valid position within its workspace. 
\textbf{Gripper Control} executes simple open-close commands with force limits to handle object grasping.

To enable robust execution, we compose these atomic controls into multi-step actions.
For example, the \textbf{Pick} primitive is decomposed into four steps:
(1) approaching a pre-grasp pose $P_{pick} + \Delta z_{pre}$;
(2) descending to $P_{pick}$;
(3) closing the gripper; and
(4) lifting to $P_{pick} + \Delta z_{post}$.
Similarly, the \textbf{Place} primitive follows a \textit{Descent–Release–Retreat} sequence to ensure stable deposition before withdrawal. Moreover, the \textbf{Go Home} primitive drives the robotic arm back to the initial position.
Beyond these basic primitives, our framework supports flexible composition of atomic actions to construct more complex assistive behaviors.
For example, we define a \textbf{Pour} primitive, in which the end-effector rotates to a specified angle to execute a pouring motion.
More sophisticated skills can be readily created by rearranging and combining existing primitives, enabling extensible assistive-arm behaviors without modifying the underlying control framework.

\subsection{Human-in-the-Loop Supervision}
\label{sec:supervision}

To ensure safety and enable flexible error recovery during long-horizon tasks, we incorporate a real-time voice supervision mechanism. This module allows the human operator to intervene during the agent’s execution. We employ Paraformer-Realtime~\cite{gao2023funasr}, an Automatic Speech Recognition (ASR) model, to transcribe operator’s voice commands. The module runs asynchronously as a background thread and continuously monitors incoming voice input regardless of the current execution stage. Once a new command is detected, the ASR output is parsed to extract key tokens, including the target arm and the intended action (e.g., ``pause'', ``retry'', or ``stop''). These tokens are then mapped to corresponding motion primitives or system-level triggers for the designated assistive arm.

Voice commands always take priority over the current execution plan. When a valid intervention is recognized, the assistive arms immediately suspend subsequent actions. This mechanism enables interactive error recovery during execution. For example, if an assistive arm fails to grasp an object, the operator can issue a ``retry'' command, causing the system to re-evaluate the current scene and re-execute the corresponding plan without disrupting the ongoing executions of other arms. In addition, a ``stop'' command can immediately halt both assistive arms, return them to their initial positions, and clear all remaining stages in the execution queue. 
In strongly coupled tasks, HATS achieves human-agent synchronization through implicit coordination and explicit voice intervention. The assistive arms execute deterministic primitives with predictable timing and motion patterns, enabling operators to adapt to the rhythm. At critical stages, voice commands further allow dynamic adjustment to maintain precise alignment. This human-in-the-loop workflow improves system safety while reducing data collection failures caused by recoverable execution errors.


\section{Experiments}
To comprehensively evaluate the proposed framework, we designed experiments to answer four research questions: 
(1) Does HATS improve data collection efficiency compared to baselines? 
(2) Does the system generate high-quality demonstrations?
(3) How do the modules achieve system efficiency and safety?
(4) Does the system reduce the labor and cognitive load of human operators?


\begin{figure}[t]
    \centering

    \begin{subfigure}[b]{0.24\textwidth}
        \centering
        \includegraphics[width=\textwidth]{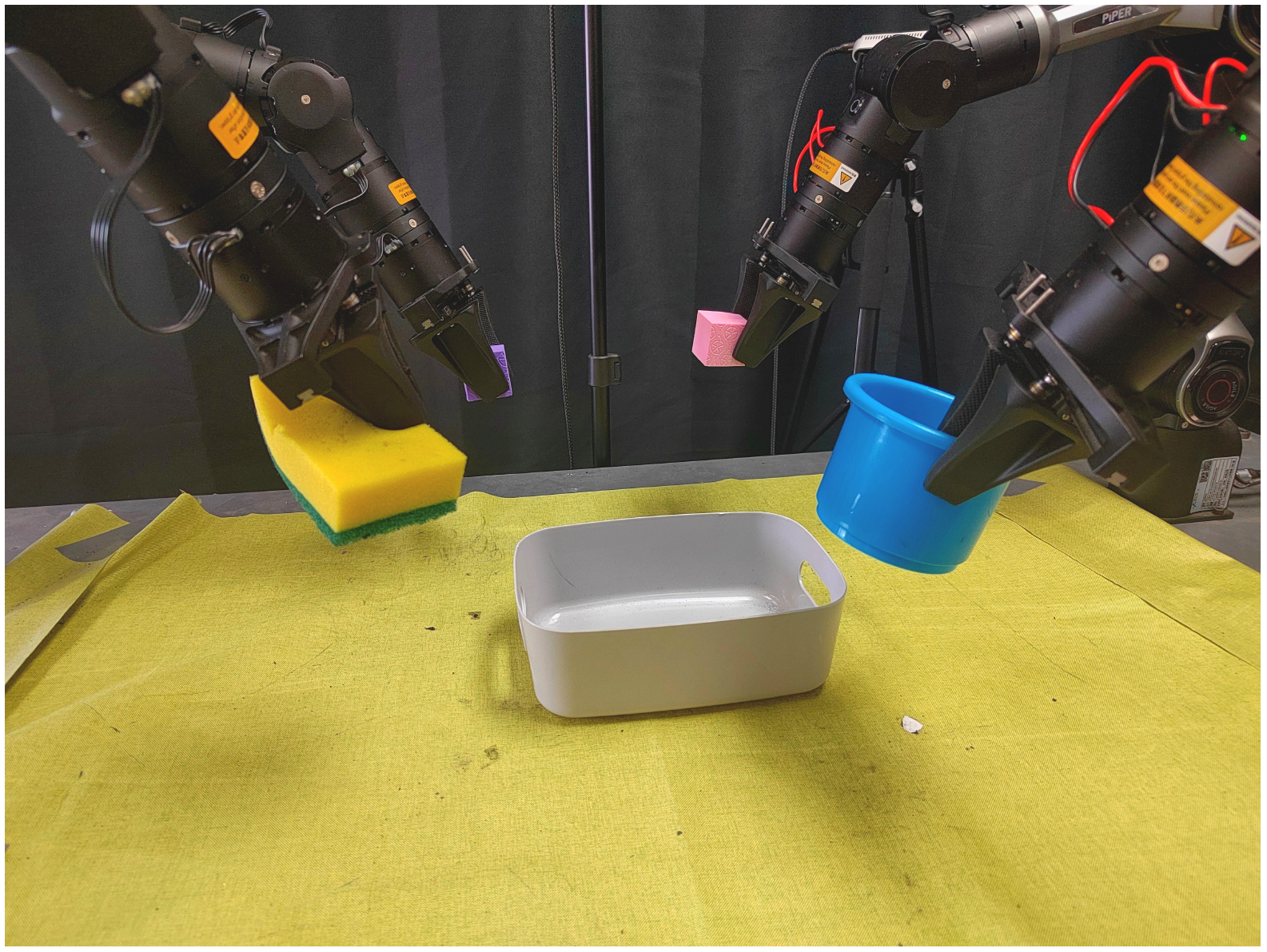}
        \vspace{-0.5cm}
        \caption{Clear}
    \end{subfigure}\hfill
    \begin{subfigure}[b]{0.24\textwidth}
        \centering
        \includegraphics[width=\textwidth]{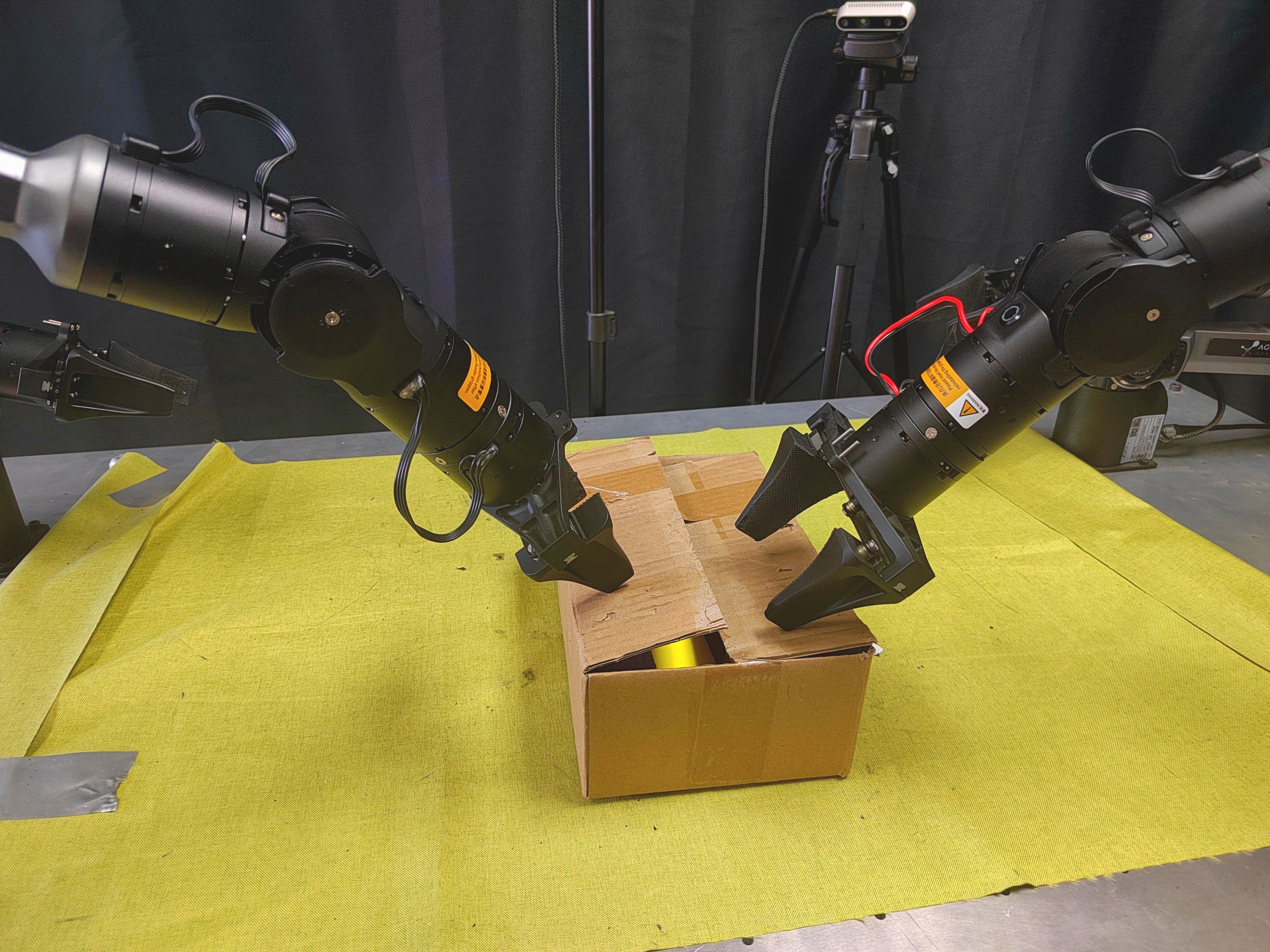}
        \vspace{-0.5cm}
        \caption{Pack}
    \end{subfigure}\hfill
    \begin{subfigure}[b]{0.24\textwidth}
        \centering
        \includegraphics[width=\textwidth]{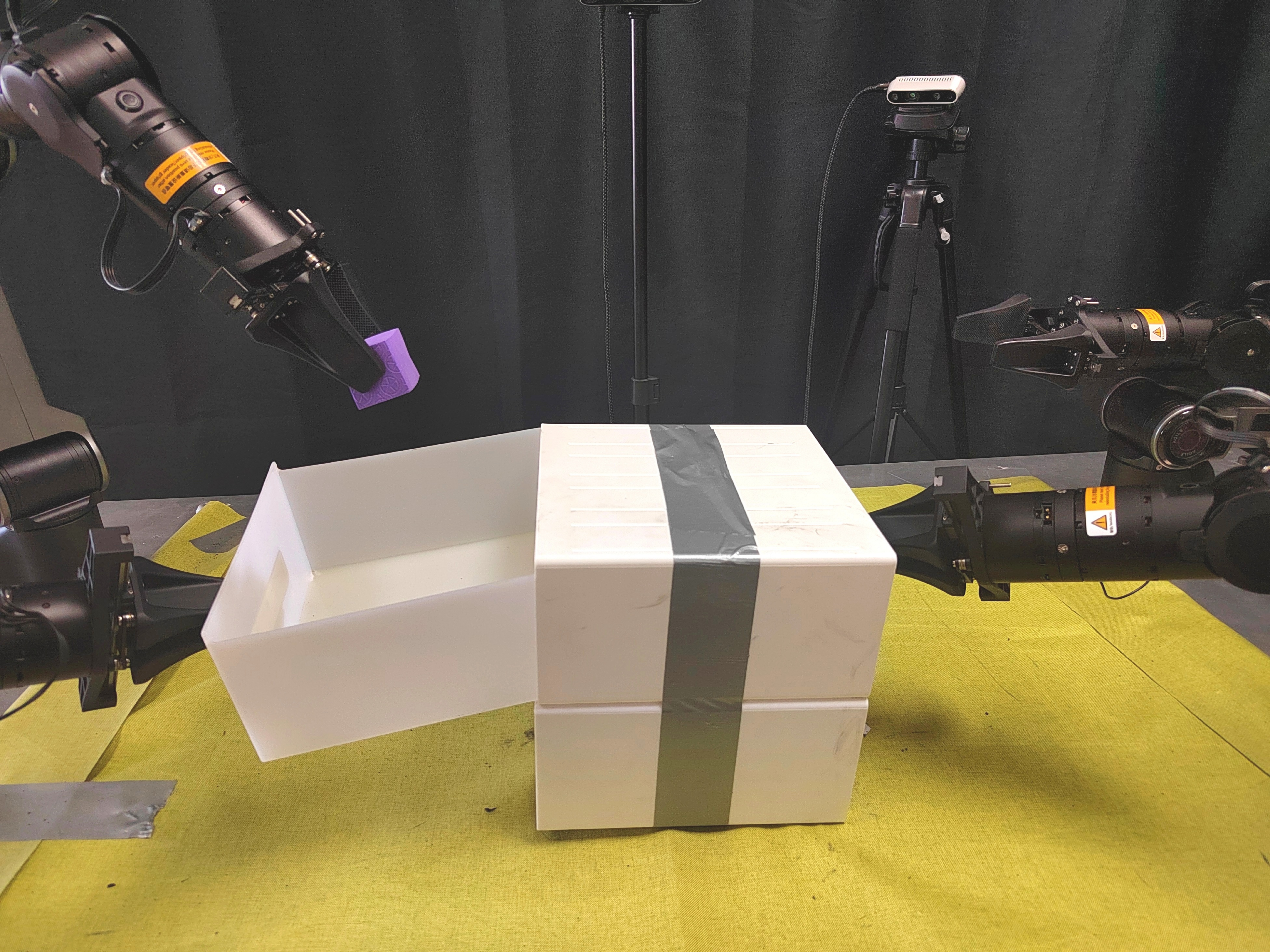}
        \vspace{-0.5cm}
        \caption{Draw}
    \end{subfigure}\hfill
    \begin{subfigure}[b]{0.24\textwidth}
        \centering
        \includegraphics[width=\textwidth]{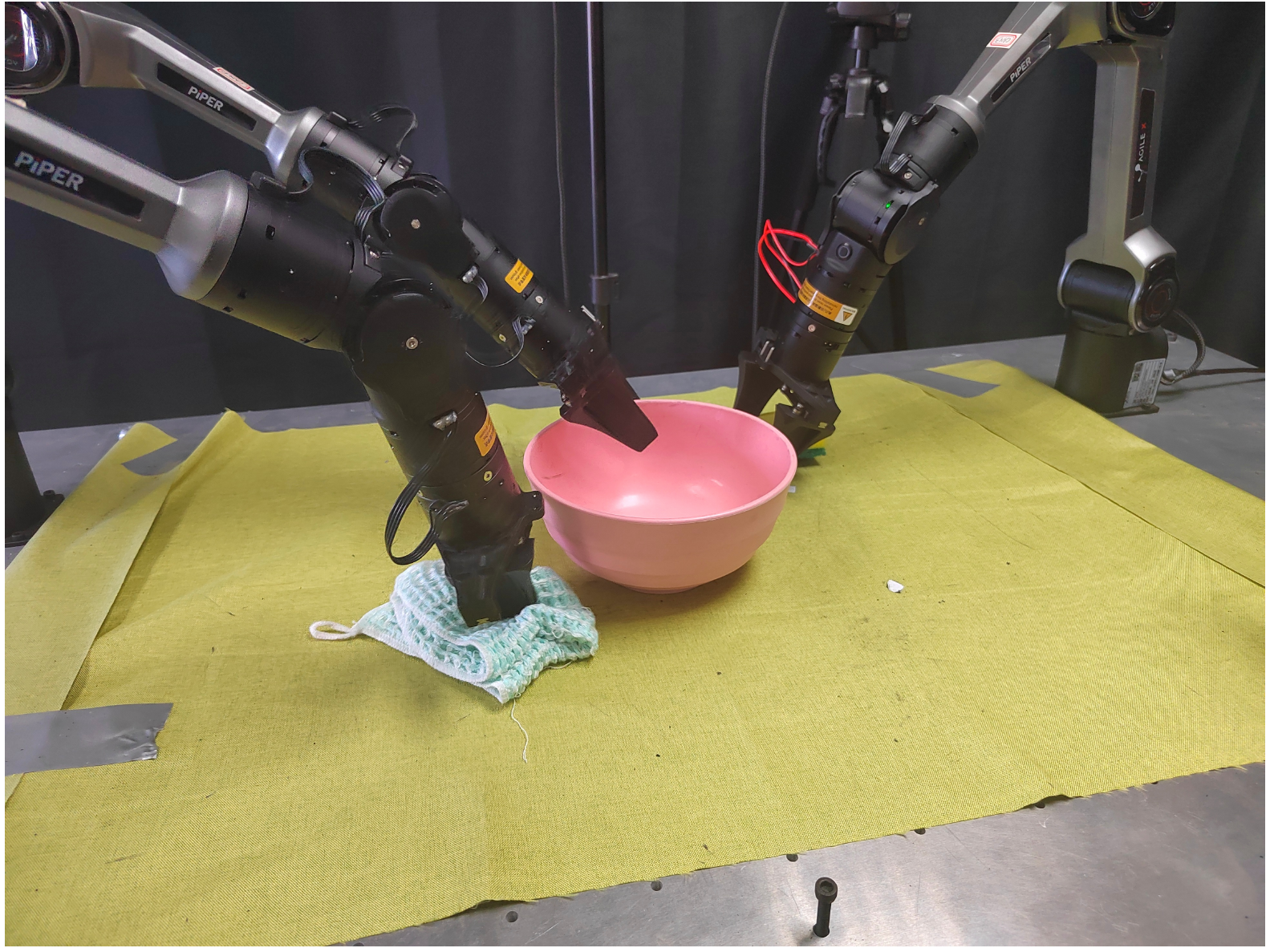}
        \vspace{-0.5cm}
        \caption{Clean}
    \end{subfigure}
    
    \vspace{0.0cm}

    \begin{subfigure}[b]{0.24\textwidth}
        \centering
        \includegraphics[width=\textwidth]{pics/Pour.pdf}
        \vspace{-0.5cm}
        \caption{Pour}
    \end{subfigure}\hspace{0.013\textwidth}%
    \begin{subfigure}[b]{0.24\textwidth}
        \centering
        \includegraphics[width=\textwidth]{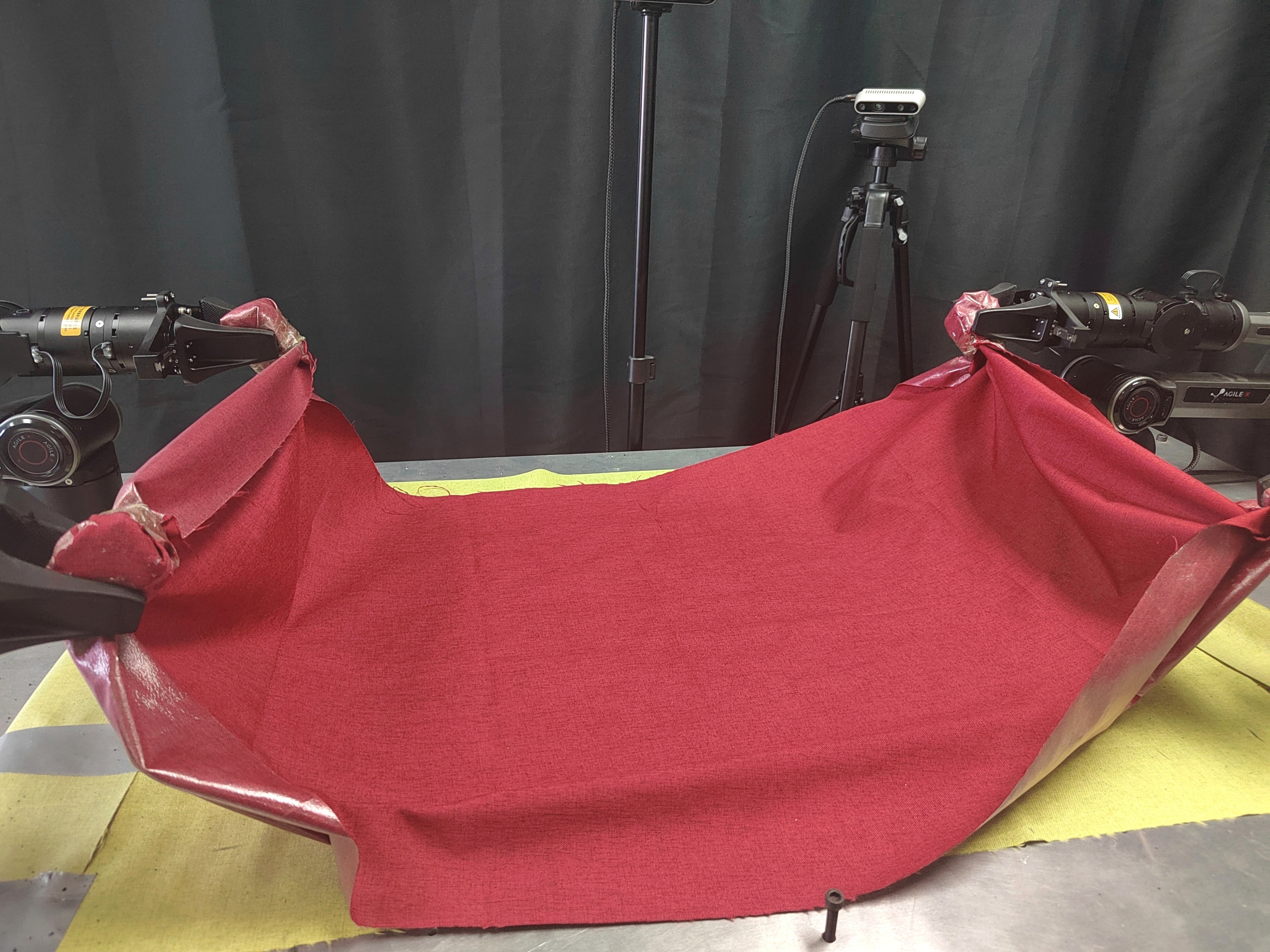}
        \vspace{-0.5cm}
        \caption{Spread}
    \end{subfigure}\hspace{0.013\textwidth}%
    \begin{subfigure}[b]{0.24\textwidth}
        \centering
        \includegraphics[width=\textwidth]{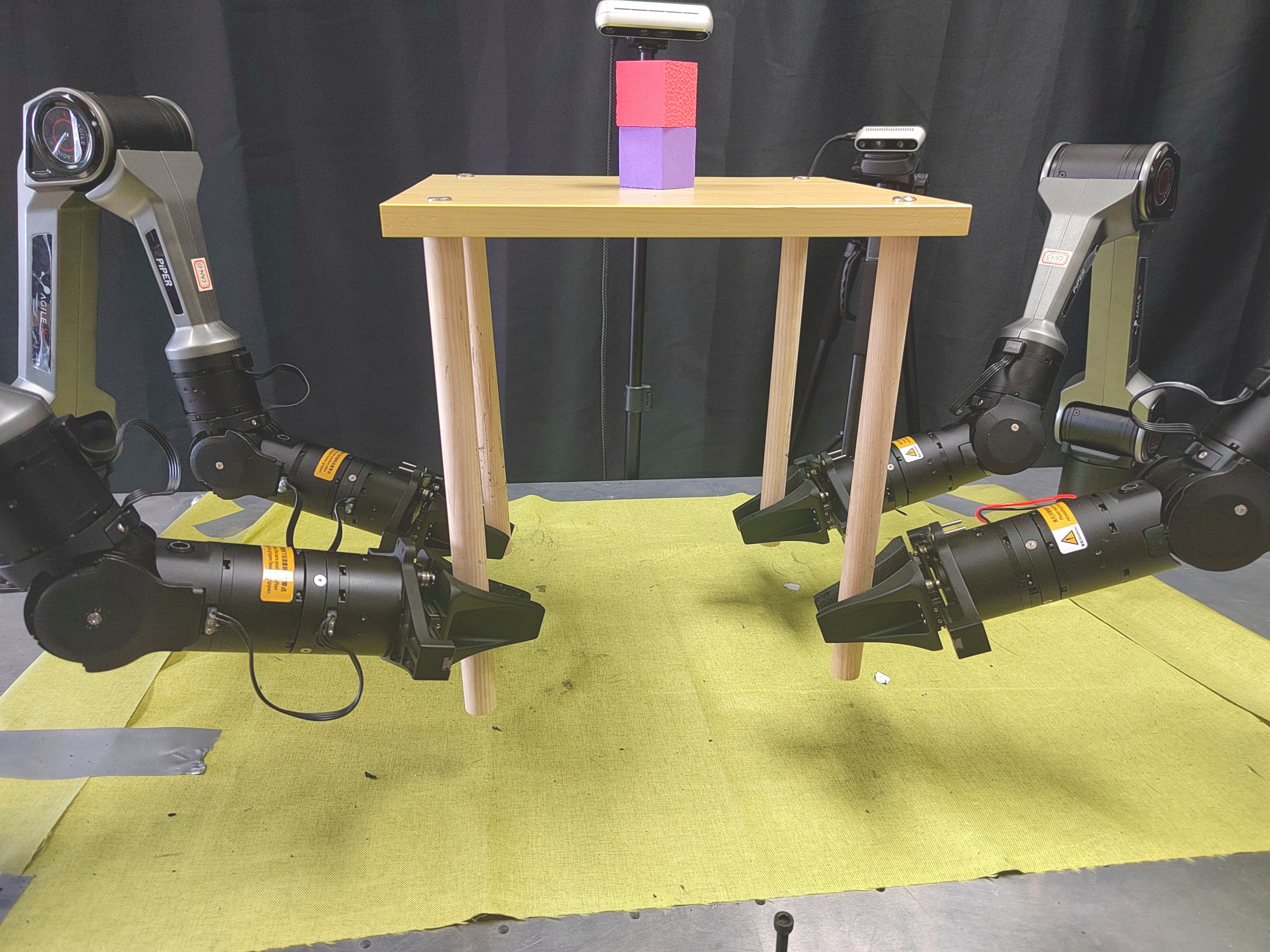}
        \vspace{-0.5cm}
        \caption{Lift}
    \end{subfigure}
    \vspace{-0.1cm}
    \caption{Visualization of the execution process for all tasks.}
    \label{fig:tasks}
    \vspace{-0.5cm}
\end{figure}

\subsection{Tasks}
As shown in  Figure ~\ref{fig:tasks}, we design seven real-world tasks grouped into three levels according to the degree of temporal and physical coupling among the four arms. 
\textbf{Independent Execution} (\textit{Clear}): the arms perform spatially independent subtasks without explicit synchronization, evaluating whether four-arm teleoperation improves efficiency over dual-arm setups. 
\textbf{Weakly Coupled Coordination} (\textit{Pack}, \textit{Draw}, \textit{Clean}, \textit{Pour}): the tasks introduce temporal dependencies, where failure or delay of one arm affects the others. 
\textbf{Strongly Coupled Coordination} (\textit{Spread}, \textit{Lift}): the tasks require tightly synchronized physically coupled manipulation by all four arms. Details of the tasks are provided in the appendix~\ref{sec: Details of Tasks}.


\subsection{System Evaluation}
We evaluate our framework against three baselines:
(1) Solo-Switching: A single human operator controls the two Pilot arms and switches control to the Copilot arms when necessary.
(2) Full-Agent: A single agent autonomously controls all four arms via the proposed HATS architecture without human involvement.
(3) Dual-Human: Two expert operators collaborate, each controlling a pair of arms simultaneously. This represents the theoretical performance ceiling. 
In addition, Table~\ref{tab:teleop_results} summarizes the Success Rate (SR) over 20 trials and the average single-trial Completion Time (CT) for each task.

\begin{table}[t]
  \centering
  \caption{\textbf{System Evaluation: HATS vs. Baselines. }Dual-Human serves as the reference upper bound. Note that ``0/20'' indicates tasks infeasible for the corresponding system. Completion Time (CT) is averaged only over successful trials, while ``--'' denotes non-applicable timings.}
  \vspace{0.1cm}
  \label{tab:teleop_results}
  \renewcommand{\arraystretch}{0.95} 
  \setlength{\tabcolsep}{4pt}        
  \small                             
  \begin{tabular}{cc ccccccc c} 
    \toprule
    \multirow{2}{*}{\textbf{System}} & \multirow{2}{*}{\textbf{Metric}} & \multicolumn{7}{c}{\textbf{Task}} & \multirow{2}{*}{\textbf{Avg.}} \\
    \cmidrule(lr){3-9}
    & & \textbf{Clear} & \textbf{Pack} & \textbf{Draw} & \textbf{Clean} & \textbf{Pour} & \textbf{Spread} & \textbf{Lift} & \\
    \midrule
    
    \multirow{2}{*}{Solo-Switching} 
      & SR $\uparrow$ & 18/20 & 19/20 & 16/20 & \textbf{18/20} & 0/20 & 0/20 & 0/20 & 50.7\% \\
      & CT (s) $\downarrow$ & 16.57 & 13.88 & 14.01 & 16.77 & -- & -- & -- & 15.31 \\
    \midrule

    \multirow{2}{*}{Full-Agent} 
      & SR  & 17/20 & 0/20 & 0/20 & 0/20 & 0/20 & 14/20 & \textbf{18/20} & 35.0\% \\
      & CT (s)  & 12.62 & -- & -- & -- & -- & \textbf{6.09} & \textbf{4.28} & \textbf{7.66} \\
    \midrule
    
    \multirow{2}{*}{\textbf{Ours (HATS)}} 
      & SR  & \textbf{18/20} & \textbf{19/20} & \textbf{19/20} & 17/20 & \textbf{14/20} & \textbf{16/20} & \textbf{18/20} & \textbf{86.4\%} \\
      & CT (s)  & \textbf{11.46} & \textbf{11.57} & \textbf{9.94} & \textbf{12.39} & \textbf{14.07} & 6.86 & 5.45 & 10.25 \\
    \midrule
    
    \multirow{2}{*}{\textcolor{darkgray}{\textit{Dual-Human}}} 
      & \textcolor{gray}{SR} & \textcolor{gray}{19/20} & \textcolor{gray}{20/20} & \textcolor{gray}{19/20} & \textcolor{gray}{20/20} & \textcolor{gray}{15/20} & \textcolor{gray}{17/20} & \textcolor{gray}{16/20} & \textcolor{gray}{90.0\%} \\
      & \textcolor{gray}{CT (s)} & \textcolor{gray}{11.39} & \textcolor{gray}{10.87} & \textcolor{gray}{9.15} & \textcolor{gray}{12.05} & \textcolor{gray}{12.29} & \textcolor{gray}{7.73} & \textcolor{gray}{6.19} & \textcolor{gray}{9.95} \\

    \bottomrule
  \end{tabular}
  \vspace{-0.45cm}
\end{table}

\textbf{HATS optimizes temporal efficiency.} 
In decoupled tasks (\textit{Clear}, \textit{Pack}, \textit{Draw}, \textit{Clean}), HATS eliminates the substantial mode-switching pauses inherent to Solo-Switching through parallel execution. Moreover, in strongly coupled tasks (\textit{Spread}, \textit{Lift}), HATS even outperforms the Dual-Human upper bound. We attribute this to the elimination of interpersonal communication latency.

\textbf{HATS overcomes the concurrency bottleneck in strongly coupled tasks.} 
For tasks requiring simultaneous multi-arm coordination (Spread, Lift), Solo-Switching fails entirely due to the lack of concurrent control. Conversely, HATS successfully completes these tasks with success rates comparable to Dual-Human baselines (e.g., 18/20 in \textit{Lift}). This proves that the assistive copilot can effectively compensate for the missing degrees of freedom of a single human operator.

\textbf{Human involvement remains indispensable.} 
While the training-free Full-Agent succeeds on structurally simple tasks, it fails on complex manipulations such as \textit{Pack} and \textit{Pour}, highlighting that human involvement remains essential for collecting high-quality demonstrations, which are important for training scalable and generalizable multi-arm policies.

\begin{table}[t]
  \centering
  \footnotesize 
  \caption{\textbf{Evaluation of downstream policies.} Results are reported as \textbf{SR/TCR} (\%). We benchmark Centralized policies trained on Full-Agent, Dual-Human, and our HATS data. Furthermore, we demonstrate a highly effective \textbf{Hybrid} deployment of our HATS framework (learned policy for primary arms + primitives for assistive arms). } 
  \label{tab:policy_results}
  \setlength{\tabcolsep}{3pt} 
  \renewcommand{\arraystretch}{0.95} 
  \vspace{0.1cm}
  \begin{tabular}{cc ccccccc c} 
    \toprule
    \multirow{2}{*}{\textbf{Policy}} & \multirow{2}{*}{\textbf{Setting}} & \multicolumn{7}{c}{\textbf{Task}} & \multirow{2}{*}{\textbf{Avg.}} \\
    \cmidrule(lr){3-9}
    & & \textbf{Clear} & \textbf{Pack} & \textbf{Draw} & \textbf{Clean} & \textbf{Pour} & \textbf{Spread} & \textbf{Lift} & \\
    \midrule
    
    \multirow{4}{*}{\textbf{DP} \cite{chi2025diffusion}} 
      & Full-Agent & 5/64.4 & 0/0.0 & 0/0.0 & 0/0.0 & 0/0.0 & 5/61.3 & 5/83.1 & 2.1/29.8 \\
      & Dual-Human & 5/58.1 & 5/78.8 & 0/65.0 & 5/71.7 & 0/73.1 & 5/58.8 & 10/86.3 & 4.3/70.3 \\
      \cmidrule{2-10}
      & \textbf{Ours (Cen.)} & 5/59.4 & 5/78.8 & 5/68.3 & 5/74.2 & 5/76.3  & 5/58.8 & 5/83.0 & 5.0/71.3 \\
      & \textbf{Ours (Hyb.)} & \textbf{5}/\textbf{77.5} & \textbf{85}/\textbf{97.8} & \textbf{15}/\textbf{84.2} & \textbf{65}/\textbf{92.5} & \textbf{70}/\textbf{94.6} & \textbf{5}/\textbf{73.1} & \textbf{20}/\textbf{88.4} & \textbf{37.9}/\textbf{86.9} \\
    \midrule
    
    \multirow{4}{*}{\textbf{DP3} \cite{ze20243d}} 
      & Full-Agent & 10/76.9 & 0/0.0 & 0/0.0 & 0/0.0 & 0/0.0 & 5/63.1 & 5/83.1 & 2.9/31.9 \\
      & Dual-Human & 5/61.9 & 5/80.6 & 5/66.7 & 0/50.0 & 0/73.1 & 5/58.8 & 10/86.3 & 4.3/68.2 \\
      \cmidrule{2-10}
      & \textbf{Ours (Cen.)} & 5/62.5 & 5/80.6 & 0/65.0 & 5/60.0 & 5/76.3 & 5/61.9 & 5/83.8 & 4.3/70.0 \\
      & \textbf{Ours (Hyb.)} & \textbf{10}/\textbf{78.8} & \textbf{85}/\textbf{97.8} & \textbf{15}/\textbf{84.2} & \textbf{50}/\textbf{87.5} & \textbf{70}/\textbf{94.6} & \textbf{5}/\textbf{75.6} & \textbf{20}/\textbf{89.0} & \textbf{36.4}/\textbf{86.8} \\
    
    \bottomrule
  \end{tabular}
  \vspace{-0.55cm}
\end{table}

\subsection{Downstream Policy Evaluation}
To validate the utility of the collected demonstrations, we trained two classic imitation learning algorithms: Diffusion Policy (DP) \cite{chi2025diffusion} and 3D Diffusion Policy (DP3) \cite{ze20243d} with 100 demos per task. We compare centralized policies trained on data collected via HATS against those trained on baselines.
Training a centralized policy to coordinate four arms is extremely challenging, with increasing multi-agent complexity severely degrading policy generalization~\cite{qin2025robofactory}. 
We observed this exact phenomenon in our experiments, where the standard binary Success Rate (SR) became prohibitively low for all setups. Therefore, to provide a more informative analysis, we adopt the \textbf{Task Completion Rate (TCR)} \cite{yang2025robot}, which is defined as:
\begin{equation}
\text{TCR} = \frac{1}{K} \sum_{k=1}^{K} \left( \frac{1}{N} \sum_{i=1}^{N} P_{k,i} \right),
\end{equation}
where $K$ denotes the number of evaluation episodes, $N$ is the number of arms, and $P_{k,i} \in [0, 1]$ represents the fractional task completion rate for the $i$-th arm in episode $k$. Each policy was evaluated with $K=20$ rollouts per task, yielding an average 90\% confidence interval of $\pm 10\%$ for TCR.

The results are summarized in Table~\ref{tab:policy_results}. Across all data sources, the policies exhibit similar failure modes: they can reproduce coarse spatial motions but frequently fail at precise multi-arm grasping, leading to low SR despite reasonable TCR. Importantly, policies trained on HATS achieve average performance comparable to or slightly better than Dual-Human demonstrations, indicating that HATS provides high-quality training data for complex multi-arm manipulation. The Hybrid baseline achieves the highest success rates, presenting a highly pragmatic paradigm for real-world multi-arm deployment. Furthermore, The superior TCR of Full-Agent and HATS data is likely due to smoother and more consistent assistive-arm trajectories. Detailed analyses are provided in the appendix~\ref{sec:trajectory_quality_analysis}. 


\subsection{System Efficiency and Safety}
\begin{wraptable}{r}{0.38\columnwidth} 
    \vspace{-0.4cm} 
    \centering
    \footnotesize 
    \renewcommand{\arraystretch}{0.9}
    \caption{\textbf{Latency Analysis.} LLM planner runs once and the plan is cached. Execution runs in real-time, while interventions trigger immediately upon ASR detection.}
    \label{tab:latency}
    \vspace{-0.2cm}
    \setlength{\tabcolsep}{2.0pt} 
    \begin{tabular}{ccc}
        \toprule
        \textbf{Component} & \textbf{Latency} & \textbf{Frequency} \\ 
        \midrule
         AnyGrasp & 0.8 s & Per Trial \\
         VLM      & 4.5 s & Per Trial \\
         Planner      & 3.6 s & Per Task \\
         Primitives& $<$0.01 s & Real-time \\
         ASR      & 1.5 s & Event \\
        \bottomrule
    \end{tabular}
    \vspace{-0.3cm} 
\end{wraptable}

As detailed in Table~\ref{tab:latency}, planner reasoning results are cached per task for high-throughput data collection. Perception is executed before each trial, enabling fluid and uninterrupted teleoperation via primitives. An asynchronous ASR thread enables rapid human-in-the-loop intervention for safety-critical halts and online error recovery. In collision stress tests, HATS achieves a 90\% intervention success rate, with the only failure caused by an ASR transcription error, whereas the unsupervised baseline fails in all cases. A physical emergency stop serves as a final safety fallback.


\subsection{User Study}

\begin{wrapfigure}{r}{0.45\columnwidth} 
    \vspace{-0.4cm} 
    \centering
    \includegraphics[width=\linewidth]{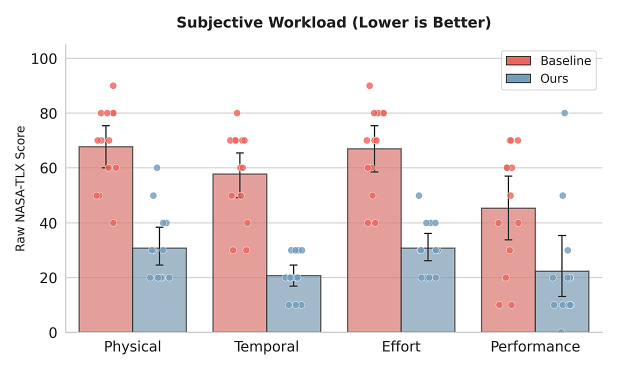}
    \caption{\textbf{Subjective Workload.} NASA-TLX results (lower is better) show HATS reduces physical and temporal demands.}
    \label{fig:nasa_tlx}
    \vspace{-0.3cm} 
\end{wrapfigure}

To evaluate system usability, we recruited 13 novice participants (each receiving a 10-minute training) to conduct two comparative experiments. We benchmarked HATS against the Dual-Human setup across all tasks. HATS achieved an average SR of 86.58\%, highly comparable to the Dual-Human baseline (89.42\%). This demonstrates that our framework effectively compensates for the absence of a second operator, substantially lowering the barrier to multi-arm teleoperation.
To quantify cognitive and physical benefits, subjective workload was measured using a simplified NASA-TLX \cite{hart1988development}. As shown in Figure ~\ref{fig:nasa_tlx}, evaluations reveal that HATS significantly reduces both physical and temporal demands against Solo-Switching. User feedback confirmed that delegating assistive arms to the agent minimizes overall workload.

\section{Conclusion}
\label{sec:conclusion}
In this work, we present \textbf{HATS}, a human–agent teleoperation system that enables a single operator assisted by an MLLM-based agent to effectively control a four-arm robotic system. HATS decouples high-level semantic planning from low-level motion execution, allowing novice users to achieve task success rates comparable to expert dual-human teams while significantly reducing cognitive and physical workload. Extensive evaluations across seven tasks further show that HATS produces high-quality demonstrations, which serve as effective training data for multi-arm manipulation policies.


\section{Limitations and Future Works}
To ensure execution stability and high task success rates during multi-arm teleoperation, HATS currently relies on structured task decomposition, predefined motion primitives, and explicit voice supervision. While this design guarantees robust and predictable human-agent coordination, it inherently trades off rapid adaptability in highly dynamic manipulation scenarios and the ability to proactively infer human intent. To achieve more fluid and tightly coupled collaboration, future work will integrate World Action Models and learned coordination policies.


\bibliography{main}  

\begin{thebibliography}{45}
\providecommand{\natexlab}[1]{#1}
\providecommand{\url}[1]{\texttt{#1}}
\expandafter\ifx\csname urlstyle\endcsname\relax
  \providecommand{\doi}[1]{doi: #1}\else
  \providecommand{\doi}{doi: \begingroup \urlstyle{rm}\Url}\fi

\bibitem[Jang et~al.(2022)Jang, Irpan, Khansari, Kappler, Ebert, Lynch, Levine, and Finn]{jang2022bc}
E.~Jang, A.~Irpan, M.~Khansari, D.~Kappler, F.~Ebert, C.~Lynch, S.~Levine, and C.~Finn.
\newblock Bc-z: Zero-shot task generalization with robotic imitation learning.
\newblock In \emph{CoRL}, 2022.
\newblock \url{https://proceedings.mlr.press/v164/jang22a.html}.

\bibitem[Belkhale et~al.(2023)Belkhale, Cui, and Sadigh]{belkhale2023data}
S.~Belkhale, Y.~Cui, and D.~Sadigh.
\newblock Data quality in imitation learning.
\newblock In \emph{NeurIPS}, 2023.
\newblock \url{https://arxiv.org/abs/2306.02437}.

\bibitem[Qin et~al.(2023)Qin, Yang, Huang, Van~Wyk, Su, Wang, Chao, and Fox]{qin2023anyteleop}
Y.~Qin, W.~Yang, B.~Huang, K.~Van~Wyk, H.~Su, X.~Wang, Y.-W. Chao, and D.~Fox.
\newblock Anyteleop: A general vision-based dexterous robot arm-hand teleoperation system.
\newblock In \emph{RSS}, 2023.
\newblock \url{https://www.roboticsproceedings.org/rss19/p015.html}.

\bibitem[Wu et~al.(2024)Wu, Shentu, Yi, Lin, and Abbeel]{wu2024gello}
P.~Wu, Y.~Shentu, Z.~Yi, X.~Lin, and P.~Abbeel.
\newblock Gello: A general, low-cost, and intuitive teleoperation framework for robot manipulators.
\newblock In \emph{IROS}, 2024.
\newblock \url{https://ieeexplore.ieee.org/document/10801581}.

\bibitem[Fang et~al.(2024)Fang, Fang, Wang, Ren, Chen, Zhang, Wang, and Lu]{fang2024airexo}
H.~Fang, H.-S. Fang, Y.~Wang, J.~Ren, J.~Chen, R.~Zhang, W.~Wang, and C.~Lu.
\newblock Airexo: Low-cost exoskeletons for learning whole-arm manipulation in the wild.
\newblock In \emph{ICRA}, 2024.
\newblock \url{https://ieeexplore.ieee.org/document/10610799}.

\bibitem[Chi et~al.(2024)Chi, Xu, Pan, Cousineau, Burchfiel, Feng, Tedrake, and Song]{chi2024universal}
C.~Chi, Z.~Xu, C.~Pan, E.~Cousineau, B.~Burchfiel, S.~Feng, R.~Tedrake, and S.~Song.
\newblock Universal manipulation interface: In-the-wild robot teaching without in-the-wild robots.
\newblock In \emph{RSS}, 2024.
\newblock \url{https://www.roboticsproceedings.org/rss20/p045.html}.

\bibitem[Liu et~al.(2025)Liu, Li, Shaw, Tao, Salakhutdinov, and Pathak]{liu2025factr}
J.~J. Liu, Y.~Li, K.~Shaw, T.~Tao, R.~Salakhutdinov, and D.~Pathak.
\newblock Factr: Force-attending curriculum training for contact-rich policy learning.
\newblock In \emph{RSS}, 2025.
\newblock \url{https://www.roboticsproceedings.org/rss21/p079.html}.

\bibitem[Tung et~al.(2021)Tung, Wong, Mandlekar, Mart{\'\i}n-Mart{\'\i}n, Zhu, Fei-Fei, and Savarese]{tung2021learning}
A.~Tung, J.~Wong, A.~Mandlekar, R.~Mart{\'\i}n-Mart{\'\i}n, Y.~Zhu, L.~Fei-Fei, and S.~Savarese.
\newblock Learning multi-arm manipulation through collaborative teleoperation.
\newblock In \emph{ICRA}, 2021.
\newblock \url{https://ieeexplore.ieee.org/document/9561491}.

\bibitem[Ozdamar et~al.(2022)Ozdamar, Laghi, Grioli, Ajoudani, Catalano, and Bicchi]{ozdamar2022shared}
I.~Ozdamar, M.~Laghi, G.~Grioli, A.~Ajoudani, M.~G. Catalano, and A.~Bicchi.
\newblock A shared autonomy reconfigurable control framework for telemanipulation of multi-arm systems.
\newblock \emph{IEEE RA-L}, 2022.
\newblock \url{https://ieeexplore.ieee.org/document/9830842}.

\bibitem[Kennel-Maushart et~al.(2021)Kennel-Maushart, Poranne, and Coros]{kennel2021manipulability}
F.~Kennel-Maushart, R.~Poranne, and S.~Coros.
\newblock Manipulability optimization for multi-arm teleoperation.
\newblock In \emph{ICRA}, 2021.
\newblock \url{https://ieeexplore.ieee.org/document/9561105}.

\bibitem[Garate et~al.(2021)Garate, Gholami, and Ajoudani]{garate2021scalable}
V.~R. Garate, S.~Gholami, and A.~Ajoudani.
\newblock A scalable framework for multi-robot tele-impedance control.
\newblock \emph{IEEE TRO}, 2021.
\newblock \url{https://ieeexplore.ieee.org/document/9429911}.

\bibitem[Kennel-Maushart et~al.(2022)Kennel-Maushart, Poranne, and Coros]{kennel2022multi}
F.~Kennel-Maushart, R.~Poranne, and S.~Coros.
\newblock Multi-arm payload manipulation via mixed reality.
\newblock In \emph{ICRA}, 2022.
\newblock \url{https://ieeexplore.ieee.org/document/9811580}.

\bibitem[Huang et~al.(2020)Huang, Eden, Cao, Burdet, and Phee]{huang2020tri}
Y.~Huang, J.~Eden, L.~Cao, E.~Burdet, and S.~J. Phee.
\newblock Tri-manipulation: An evaluation of human performance in 3-handed teleoperation.
\newblock \emph{IEEE Transactions on Medical Robotics and Bionics}, 2020.
\newblock \url{https://ieeexplore.ieee.org/document/9235524}.

\bibitem[Mandlekar et~al.(2021)Mandlekar, Xu, Wong, Nasiriany, Wang, Kulkarni, Fei-Fei, Savarese, Zhu, and Mart{\'\i}n-Mart{\'\i}n]{mandlekar2021matters}
A.~Mandlekar, D.~Xu, J.~Wong, S.~Nasiriany, C.~Wang, R.~Kulkarni, L.~Fei-Fei, S.~Savarese, Y.~Zhu, and R.~Mart{\'\i}n-Mart{\'\i}n.
\newblock What matters in learning from offline human demonstrations for robot manipulation.
\newblock In \emph{CoRL}, 2021.
\newblock \url{https://openreview.net/forum?id=JrsfBJtDFdI}.

\bibitem[Fu et~al.(2024)Fu, Zhao, and Finn]{fu2024mobile}
Z.~Fu, T.~Z. Zhao, and C.~Finn.
\newblock Mobile aloha: Learning bimanual mobile manipulation with low-cost whole-body teleoperation.
\newblock \emph{arXiv preprint arXiv:2401.02117}, 2024.
\newblock \url{https://arxiv.org/abs/2401.02117}.

\bibitem[Zhao et~al.(2023)Zhao, Kumar, Levine, and Finn]{zhao2023learning}
T.~Z. Zhao, V.~Kumar, S.~Levine, and C.~Finn.
\newblock Learning fine-grained bimanual manipulation with low-cost hardware.
\newblock In \emph{RSS}, 2023.
\newblock \url{https://www.roboticsproceedings.org/rss19/p016.html}.

\bibitem[Aldaco et~al.(2024)Aldaco, Armstrong, Baruch, Bingham, Chan, Draper, Dwibedi, Finn, Florence, Goodrich, et~al.]{aldaco2024aloha}
J.~Aldaco, T.~Armstrong, R.~Baruch, J.~Bingham, S.~Chan, K.~Draper, D.~Dwibedi, C.~Finn, P.~Florence, S.~Goodrich, et~al.
\newblock Aloha 2: An enhanced low-cost hardware for bimanual teleoperation.
\newblock \emph{arXiv preprint arXiv:2405.02292}, 2024.
\newblock \url{https://arxiv.org/abs/2405.02292}.

\bibitem[Handa et~al.(2020)Handa, Van~Wyk, Yang, Liang, Chao, Wan, Birchfield, Ratliff, and Fox]{handa2020dexpilot}
A.~Handa, K.~Van~Wyk, W.~Yang, J.~Liang, Y.-W. Chao, Q.~Wan, S.~Birchfield, N.~Ratliff, and D.~Fox.
\newblock Dexpilot: Vision-based teleoperation of dexterous robotic hand-arm system.
\newblock In \emph{ICRA}, 2020.
\newblock \url{https://ieeexplore.ieee.org/document/9197124}.

\bibitem[Chen et~al.(2025)Chen, Wang, Nguyen, Fei-Fei, and Liu]{chen2025arcap}
S.~Chen, C.~Wang, K.~Nguyen, L.~Fei-Fei, and C.~K. Liu.
\newblock Arcap: Collecting high-quality human demonstrations for robot learning with augmented reality feedback.
\newblock In \emph{ICRA}, 2025.
\newblock \url{https://ieeexplore.ieee.org/document/11128717/}.

\bibitem[Laghi et~al.(2018)Laghi, Maimeri, Marchand, Leparoux, Catalano, Ajoudani, and Bicchi]{laghi2018shared}
M.~Laghi, M.~Maimeri, M.~Marchand, C.~Leparoux, M.~Catalano, A.~Ajoudani, and A.~Bicchi.
\newblock Shared-autonomy control for intuitive bimanual tele-manipulation.
\newblock In \emph{IEEE-RAS Humanoids}, 2018.
\newblock \url{https://ieeexplore.ieee.org/document/8625047}.

\bibitem[Manschitz and Ruiken(2022)]{manschitz2022shared}
S.~Manschitz and D.~Ruiken.
\newblock Shared autonomy for intuitive teleoperation.
\newblock In \emph{ICRA Workshop}, 2022.
\newblock \url{https://www.honda-ri.de/pubs/pdf/5098.pdf}.

\bibitem[Fu et~al.(2025)Fu, Song, Hu, and Detry]{fu2025tasc}
Z.~Fu, P.~Song, Y.~Hu, and R.~Detry.
\newblock Tasc: Task-aware shared control for teleoperated manipulation.
\newblock \emph{arXiv preprint arXiv:2509.10416}, 2025.
\newblock \url{https://arxiv.org/abs/2509.10416}.

\bibitem[Morrell et~al.(2021)Morrell, Morrell-Junior, Morrell, Mendes, MAURICIO, TUSTUMI, Morrell, et~al.]{morrell2021history}
A.~L.~G. Morrell, A.~C. Morrell-Junior, A.~G. Morrell, J.~Mendes, F.~MAURICIO, F.~TUSTUMI, A.~Morrell, et~al.
\newblock The history of robotic surgery and its evolution: when illusion becomes reality.
\newblock \emph{Revista do Col{\'e}gio Brasileiro de Cirurgi{\~o}es}, 2021.
\newblock \url{https://www.researchgate.net/publication/348585670_The_history_of_robotic_surgery_and_its_evolution_when_illusion_becomes_reality}.

\bibitem[Kelly et~al.(2019)Kelly, Sidrane, Driggs-Campbell, and Kochenderfer]{kelly2019hg}
M.~Kelly, C.~Sidrane, K.~Driggs-Campbell, and M.~J. Kochenderfer.
\newblock Hg-dagger: Interactive imitation learning with human experts.
\newblock In \emph{ICRA}, 2019.
\newblock \url{https://ieeexplore.ieee.org/document/8793698}.

\bibitem[Mandlekar et~al.(2020)Mandlekar, Xu, Mart{\'\i}n-Mart{\'\i}n, Zhu, Fei-Fei, and Savarese]{mandlekar2020human}
A.~Mandlekar, D.~Xu, R.~Mart{\'\i}n-Mart{\'\i}n, Y.~Zhu, L.~Fei-Fei, and S.~Savarese.
\newblock Human-in-the-loop imitation learning using remote teleoperation.
\newblock \emph{arXiv preprint arXiv:2012.06733}, 2020.
\newblock \url{https://arxiv.org/abs/2012.06733}.

\bibitem[Wu et~al.(2025)Wu, Shentu, Liao, Jin, Guo, Sreenath, Lin, and Abbeel]{wu2025robocopilot}
P.~Wu, Y.~Shentu, Q.~Liao, D.~Jin, M.~Guo, K.~Sreenath, X.~Lin, and P.~Abbeel.
\newblock Robocopilot: Human-in-the-loop interactive imitation learning for robot manipulation.
\newblock \emph{arXiv preprint arXiv:2503.07771}, 2025.
\newblock \url{https://arxiv.org/abs/2503.07771}.

\bibitem[Xu et~al.(2025)Xu, Zhao, Wu, Liu, Ji, Che, Liu, and Tang]{xu2025hacts}
Z.~Xu, Y.~Zhao, K.~Wu, N.~Liu, J.~Ji, Z.~Che, C.~H. Liu, and J.~Tang.
\newblock Hacts: a human-as-copilot teleoperation system for robot learning.
\newblock In \emph{IROS}, 2025.
\newblock \url{https://ieeexplore.ieee.org/document/11247309}.

\bibitem[Luo et~al.(2025)Luo, Peng, Lv, Hong, Driggs-Campbell, Lu, and Li]{luo2025human}
S.~Luo, Q.~Peng, J.~Lv, K.~Hong, K.~R. Driggs-Campbell, C.~Lu, and Y.-L. Li.
\newblock Human-agent joint learning for efficient robot manipulation skill acquisition.
\newblock In \emph{ICRA}, 2025.
\newblock \url{https://ieeexplore.ieee.org/document/11127637}.

\bibitem[Belsare et~al.(2025)Belsare, Karimi, Mattson, and Brown]{belsare2025toward}
A.~Belsare, Z.~Karimi, C.~Mattson, and D.~S. Brown.
\newblock Toward zero-shot user intent recognition in shared autonomy.
\newblock In \emph{HRI}, 2025.
\newblock \url{https://ieeexplore.ieee.org/document/10973985}.

\bibitem[Huang et~al.(2023)Huang, Wang, Zhang, Li, Wu, and Fei-Fei]{huang2023voxposer}
W.~Huang, C.~Wang, R.~Zhang, Y.~Li, J.~Wu, and L.~Fei-Fei.
\newblock Voxposer: Composable 3d value maps for robotic manipulation with language models.
\newblock In \emph{CoRL}, 2023.
\newblock \url{https://openreview.net/forum?id=9_8LF30mOC}.

\bibitem[Liang et~al.(2023)Liang, Huang, Xia, Xu, Hausman, Ichter, Florence, and Zeng]{liang2022code}
J.~Liang, W.~Huang, F.~Xia, P.~Xu, K.~Hausman, B.~Ichter, P.~Florence, and A.~Zeng.
\newblock Code as policies: Language model programs for embodied control.
\newblock In \emph{ICRA}, 2023.
\newblock \url{https://ieeexplore-custom.ieee.org/document/10160591/}.

\bibitem[Yang et~al.(2025)Yang, Gu, Wen, Li, Zhao, Wang, and Liu]{yang2025maniagent}
Y.~Yang, K.~Gu, Y.~Wen, H.~Li, Y.~Zhao, T.~Wang, and X.~Liu.
\newblock Maniagent: An agentic framework for general robotic manipulation.
\newblock \emph{arXiv preprint arXiv:2510.11660}, 2025.
\newblock \url{https://arxiv.org/abs/2510.11660}.

\bibitem[Tao et~al.(2025)Tao, Yang, Ding, and Erickson]{tao2025lams}
Y.~Tao, J.~Yang, D.~Ding, and Z.~Erickson.
\newblock Lams: Llm-driven automatic mode switching for assistive teleoperation.
\newblock In \emph{HRI}, 2025.
\newblock \url{https://ieeexplore.ieee.org/document/10974127}.

\bibitem[Fang et~al.(2023)Fang, Wang, Fang, Gou, Liu, Yan, Liu, Xie, and Lu]{fang2023anygrasp}
H.-S. Fang, C.~Wang, H.~Fang, M.~Gou, J.~Liu, H.~Yan, W.~Liu, Y.~Xie, and C.~Lu.
\newblock Anygrasp: Robust and efficient grasp perception in spatial and temporal domains.
\newblock \emph{IEEE TRO}, 2023.
\newblock \url{https://ieeexplore.ieee.org/document/10167687}.

\bibitem[Ester et~al.(1996)Ester, Kriegel, Sander, Xu, et~al.]{ester1996density}
M.~Ester, H.-P. Kriegel, J.~Sander, X.~Xu, et~al.
\newblock A density-based algorithm for discovering clusters in large spatial databases with noise.
\newblock In \emph{kdd}, 1996.
\newblock \url{https://cdn.aaai.org/KDD/1996/KDD96-037.pdf}.

\bibitem[Bai et~al.(2023)Bai, Bai, Yang, Wang, Tan, Wang, Lin, Zhou, and Zhou]{bai2023qwen}
J.~Bai, S.~Bai, S.~Yang, S.~Wang, S.~Tan, P.~Wang, J.~Lin, C.~Zhou, and J.~Zhou.
\newblock Qwen-vl: A frontier large vision-language model with versatile abilities.
\newblock \emph{arXiv preprint arXiv:2308.12966}, 2023.
\newblock \url{https://arxiv.org/abs/2308.12966}.

\bibitem[Gao et~al.(2023)Gao, Li, Wang, Luo, Shi, Chen, Li, Zuo, Du, Xiao, et~al.]{gao2023funasr}
Z.~Gao, Z.~Li, J.~Wang, H.~Luo, X.~Shi, M.~Chen, Y.~Li, L.~Zuo, Z.~Du, Z.~Xiao, et~al.
\newblock Funasr: A fundamental end-to-end speech recognition toolkit.
\newblock \emph{arXiv preprint arXiv:2305.11013}, 2023.
\newblock \url{https://arxiv.org/abs/2305.11013}.

\bibitem[Chi et~al.(2023)Chi, Xu, Feng, Cousineau, Du, Burchfiel, Tedrake, and Song]{chi2025diffusion}
C.~Chi, Z.~Xu, S.~Feng, E.~Cousineau, Y.~Du, B.~Burchfiel, R.~Tedrake, and S.~Song.
\newblock Diffusion policy: Visuomotor policy learning via action diffusion.
\newblock In \emph{RSS}, 2023.
\newblock \url{https://www.roboticsproceedings.org/rss19/p026.html}.

\bibitem[Ze et~al.(2024)Ze, Zhang, Zhang, Hu, Wang, and Xu]{ze20243d}
Y.~Ze, G.~Zhang, K.~Zhang, C.~Hu, M.~Wang, and H.~Xu.
\newblock 3d diffusion policy: Generalizable visuomotor policy learning via simple 3d representations.
\newblock In \emph{RSS}, 2024.
\newblock \url{https://www.roboticsproceedings.org/rss20/p067.html}.

\bibitem[Qin et~al.(2025)Qin, Kang, Song, Yin, Liu, Liu, Zhang, and Bai]{qin2025robofactory}
Y.~Qin, L.~Kang, X.~Song, Z.~Yin, X.~Liu, X.~Liu, R.~Zhang, and L.~Bai.
\newblock Robofactory: Exploring embodied agent collaboration with compositional constraints.
\newblock In \emph{ICCV}, 2025.
\newblock \url{https://arxiv.org/abs/2503.16408}.

\bibitem[Yang et~al.(2025)Yang, Eppner, Tremblay, Fox, Birchfield, and Ramos]{yang2025robot}
X.~Yang, C.~Eppner, J.~Tremblay, D.~Fox, S.~Birchfield, and F.~Ramos.
\newblock Robot policy evaluation for sim-to-real transfer: A benchmarking perspective.
\newblock \emph{arXiv preprint arXiv:2508.11117}, 2025.
\newblock \url{https://arxiv.org/abs/2508.11117}.

\bibitem[Hart and Staveland(1988)]{hart1988development}
S.~G. Hart and L.~E. Staveland.
\newblock Development of nasa-tlx (task load index): Results of empirical and theoretical research.
\newblock In \emph{Human Mental Workload}, Advances in Psychology. North-Holland, 1988.
\newblock \url{https://www.sciencedirect.com/science/article/pii/S0166411508623869}.

\bibitem[Argall et~al.(2009)Argall, Chernova, Veloso, and Browning]{argall2009survey}
B.~D. Argall, S.~Chernova, M.~Veloso, and B.~Browning.
\newblock A survey of robot learning from demonstration.
\newblock \emph{RAS}, 2009.
\newblock \url{https://www.sciencedirect.com/science/article/pii/S0921889008001772}.

\bibitem[Ravichandar et~al.(2020)Ravichandar, Polydoros, Chernova, and Billard]{ravichandar2020recent}
H.~Ravichandar, A.~S. Polydoros, S.~Chernova, and A.~Billard.
\newblock Recent advances in robot learning from demonstration.
\newblock \emph{Annual Review of Control, Robotics, and Autonomous Systems}, 2020.
\newblock \url{https://www.annualreviews.org/content/journals/10.1146/annurev-control-100819-063206}.

\bibitem[Flash and Hogan(1985)]{flash1985coordination}
T.~Flash and N.~Hogan.
\newblock The coordination of arm movements: an experimentally confirmed mathematical model.
\newblock \emph{The Journal of Neuroscience}, 1985.
\newblock \url{https://www.jneurosci.org/content/5/7/1688}.

\end{thebibliography}
\clearpage
\appendix
\begin{center}
    \LARGE\textbf{Appendix} 
\end{center}

\section{Video Demo}
A comprehensive video demo is included in the supplementary material. The video visually outlines our research motivation and the proposed HATS methodology. Furthermore, it features extensive real-world footage, showcasing both the intuitive human-agent teleoperation process during data collection and the autonomous real-world execution of our downstream learned policies across complex multi-arm tasks.

\section{Hardware System}
\label{sec:hardware_appendix}

To facilitate reproducibility and provide deeper insights into our experimental platform, we detail the visual specifications, hardware configurations, and cost analysis of our human-agent teleoperation system.

\subsection{Robot Configuration and Teleoperation Interface}
\begin{figure}[ht]
    \centering
    \begin{subfigure}[b]{0.45\columnwidth}
        \centering
        \includegraphics[width=\linewidth]{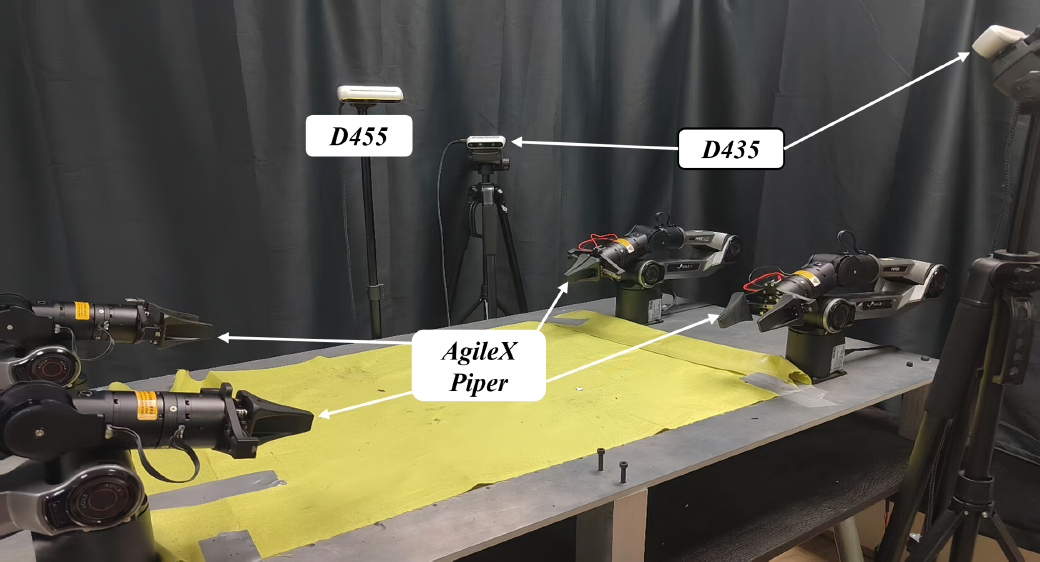}
        \caption{Physical hardware platform.}
        \label{fig:hardware_a}
    \end{subfigure}
    \hspace{2pt}
    \begin{subfigure}[b]{0.406\columnwidth}
        \centering
        \includegraphics[width=\linewidth]{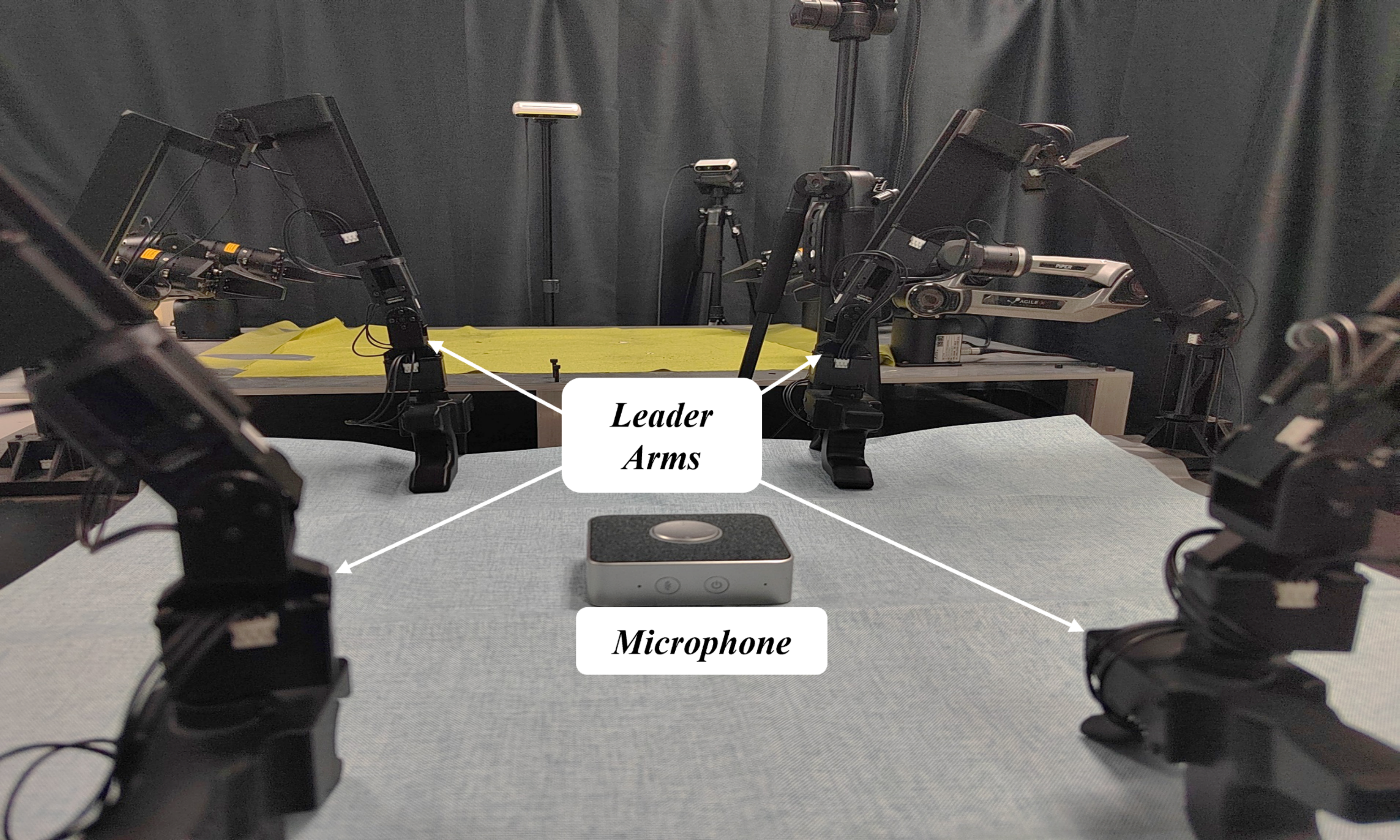}
        \caption{Teleoperation interface.}
        \label{fig:hardware_b}
    \end{subfigure}
    \vspace{-0.2cm} 
    \caption{The illustration of the hardware system setup. (a) Follower setup featuring four Agilex Piper arms and RealSense cameras. (b) The GELLO-inspired leader control interface.}
    \label{fig:hardware_setup}
    \vspace{-0.2cm}
\end{figure}

The physical hardware platform, as illustrated in Fig.~\ref{fig:hardware_setup}(a), consists of four lightweight 7-DOF Agilex Piper robotic arms. For environmental perception, the system incorporates three RGB-D cameras: an Intel RealSense D455 provides high-resolution depth data for AnyGrasp to localize object affordances, while two Intel RealSense D435 cameras capture third-person observations for downstream policy training.

For teleoperation, we adopt a Leader-Follower setup inspired by GELLO, as shown in Fig.~\ref{fig:hardware_setup}(b). In addition to the mechanical interface, a portable microphone is deployed to capture the operator’s voice commands for real-time task supervision.

\subsection{Leader Arm Design and Cost Analysis}
\begin{figure}[ht]
    \centering
    \begin{minipage}[c]{0.47\textwidth}
        \centering
        \includegraphics[width=1.0\linewidth]{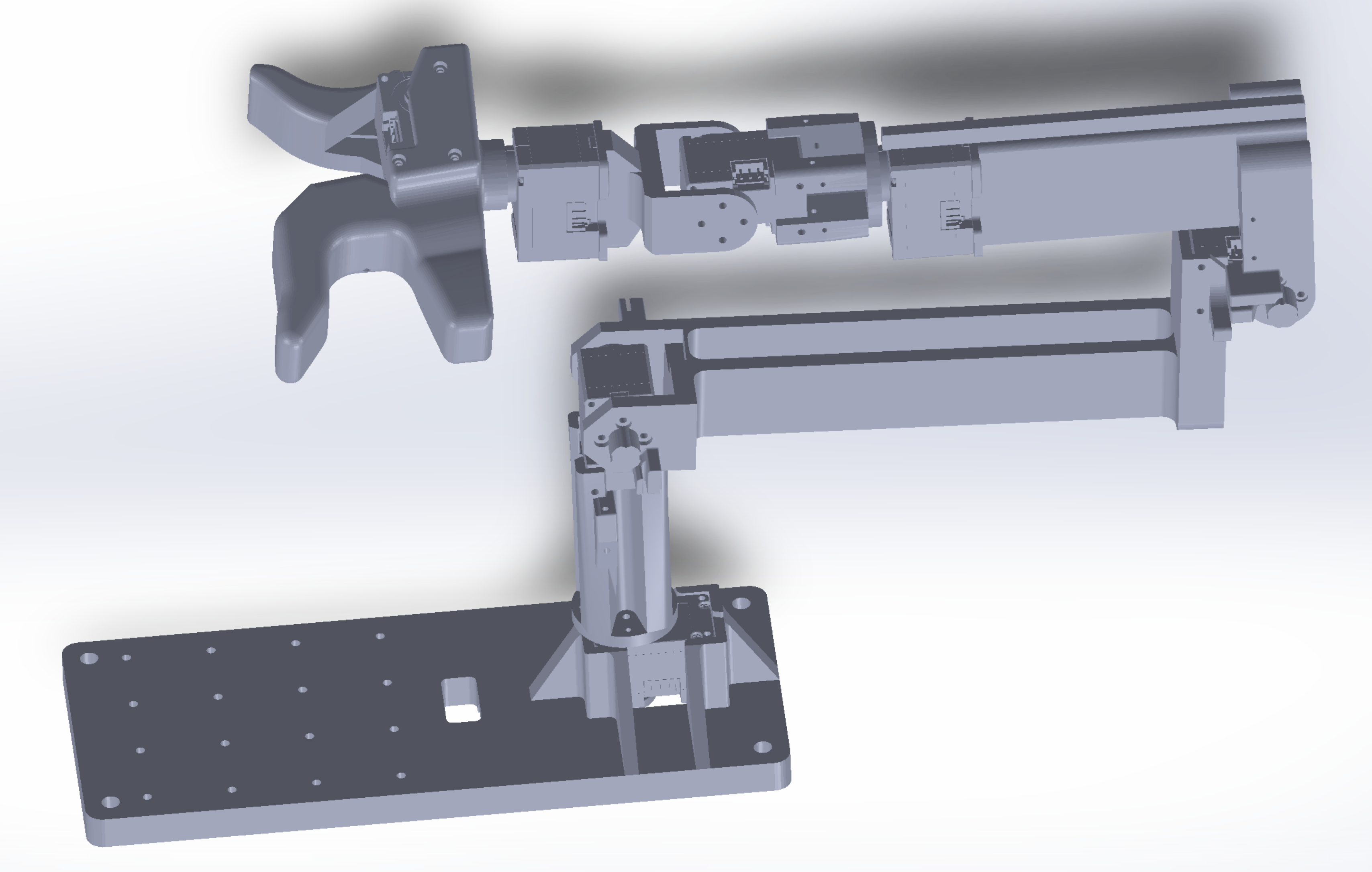}
        \captionof{figure}{3D exploded view of the custom leader arm structure.}
        \label{fig:leader_arm}
    \end{minipage}
    \hfill
    \begin{minipage}[c]{0.47\textwidth}
        \centering
        \small
        \captionof{table}{Bill of Materials (BoM) and Cost Analysis for a Single Custom Leader Arm.}
        \label{tab:cost_analysis}
        \renewcommand{\arraystretch}{1.1} 
        \begin{tabular}{l c r} 
        \toprule
        \textbf{Component} & \textbf{Qty.} & \textbf{Cost (\$)} \\
        \midrule
        Dynamixel XL330 & 7 & 235.00 \\
        U2D2 Control PCB & 1 & 66.43 \\
        U2D2 Power Hub & 1 & 33.57 \\
        FPX330-H101 Frame & 1 & 13.57 \\
        Power Supply (5V 5A) & 1 & 11.29 \\
        Cable (X3PIN 25CM) & 1 & 1.43 \\
        Fasteners \& Screws & - & 2.19 \\
        \midrule
        \textbf{Total Cost} & & \textbf{363.48} \\
        \bottomrule
        \end{tabular}
    \end{minipage}
    \vspace{-0.4cm}
\end{figure}

The leader devices are assembled using 3D-printed structural components and off-the-shelf Dynamixel servos. Crucially, they are designed to be kinematically isomorphic to the Piper follower arms. This structural parity enables direct joint-to-joint angle mapping without complex inverse kinematics (IK) retargeting, significantly improving control accuracy and reducing system latency. Fig.~\ref{fig:leader_arm} presents the 3D exploded view of the custom-built leader arm, and Table~\ref{tab:cost_analysis} details its Bill of Materials (BoM). The total cost for constructing a single leader arm is approximately \textbf{\$363.48}, making this kinematically-matched teleoperation setup a highly cost-effective solution for multi-arm research scaling.

\subsection{Workspace Layout}
\begin{figure}[h]
  \centering
  \includegraphics[width=0.9\linewidth]{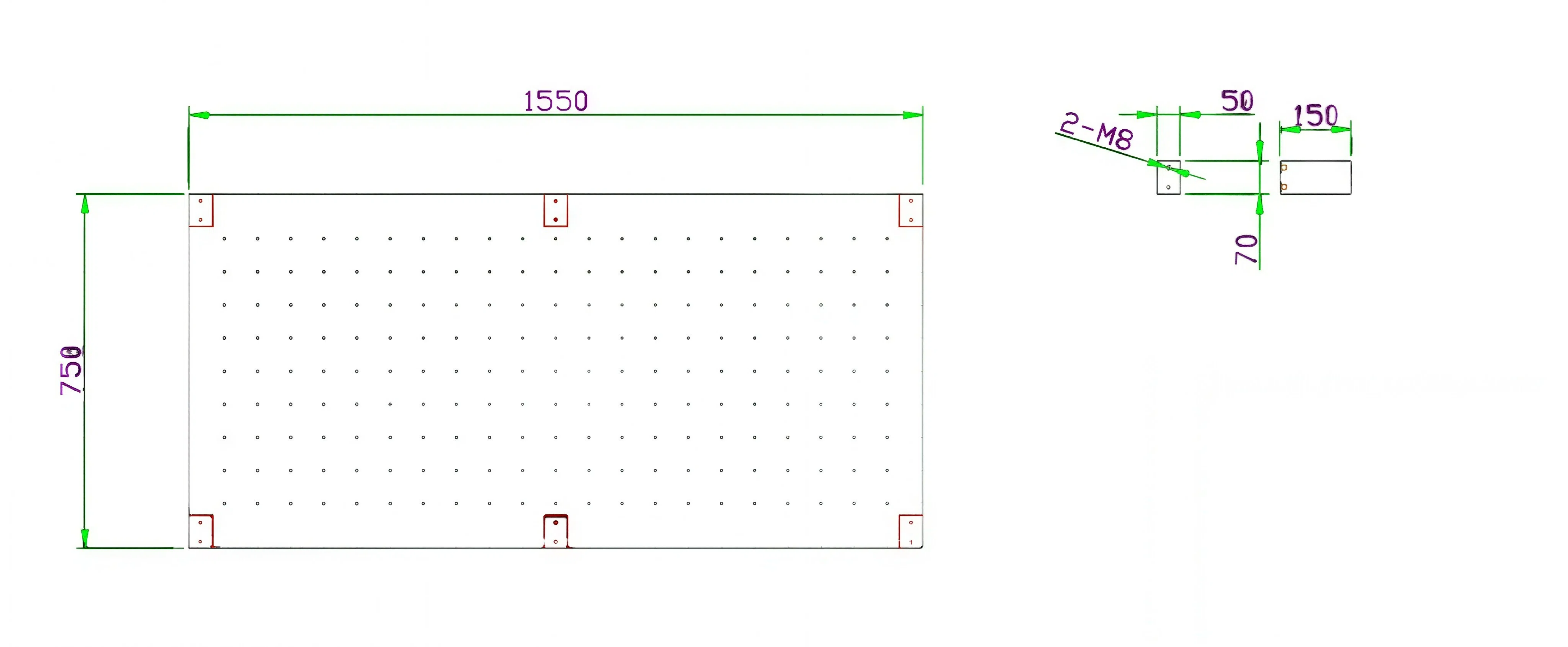} 
  \caption{Dimensions of the tabletop workspace layout.}
  \label{fig:desktop_dim}
  \vspace{-0.2cm}
\end{figure}

Finally, to ensure sufficient range of motion for four-arm coordination and collision avoidance, Fig.~\ref{fig:desktop_dim} illustrates the specific dimensions of the tabletop workspace layout used throughout our experiments.

\section{Experiments Details}
\subsection{Details of Tasks}
\label{sec: Details of Tasks}
\textit{Clear:} All four arms operate in parallel to grasp randomly scattered objects across the workspace and deposit them into a designated target bin. 

\textit{Pack:} The assistive arms execute the preparatory phase by loading items into a cardboard box. Subsequently, the primary arms sequentially fold and close the box flaps. 

\textit{Draw:} One primary arm stabilizes the fixture while the other dynamically pulls the drawer open, creating a functional workspace for an assistive arm to precisely insert an object inside. 

\textit{Clean:} An assistive arm positions a receptacle centrally, allowing a primary arm to execute a wiping motion on the adjacent table surface. Concurrently, the second assistive arm transfers scattered objects into the receptacle. 

\textit{Pour:} Two assistive arms synchronously pour granular objects from separate cups into a central container. The primary arms then collaboratively lift the container and perform a continuous shaking motion to mix the contents. 

\textit{Spread:} The four arms grasp the respective corners of a folded cloth and synchronously extend outward to flatten it against the tabletop. 

\textit{Lift:} The four arms securely grasp the edges of a small wooden table and execute a joint lifting motion, maintaining planar balance throughout the trajectory to prevent stacked blocks from toppling.

\subsection{Stage-wise TCR Calculation}
\label{sec:tcr_criteria}

As introduced in the main text, TCR assigns a normalized fractional score $P_{k,i} \in [0, 1]$ to each participating arm based on physical milestones. To ensure evaluation transparency and reproducibility, we detail the specific scoring criteria for all tasks below. 

\textbf{Clear (4 arms):} All four arms share an identical two-stage criterion: (1) successfully grasping the target object ($+0.5$), and (2) successfully releasing it into the designated bin ($+0.5$).

\textbf{Pack (4 arms):} The two assistive arms follow a standard pick-and-place criterion: grasping the item ($+0.5$) and loading it into the box ($+0.5$). The two primary arms execute a contact-rich closing motion: reaching and making contact with the box flap ($+0.5$), and successfully folding it into the closed position ($+0.5$).

\textbf{Draw (3 arms):} The single assistive arm earns credit for grasping the object ($+0.5$) and successfully placing it inside the opened drawer ($+0.5$). The active primary arm is scored for grasping the drawer handle ($+0.5$) and successfully pulling it open ($+0.5$). The stabilizing primary arm is scored for grasping the fixture handle ($+0.5$) and maintaining a stable grip to secure the drawer structure during the pulling phase ($+0.5$).

\textbf{Clean (3 arms):} The two assistive arms follow the two-stage criterion: grasping their respective targets (bowl or leftovers) ($+0.5$) and placing them in the correct location ($+0.5$). The single primary arm involves a three-stage continuous sequence: grasping the towel ($+1/3$), executing the wiping motion on the table ($+1/3$), and placing the towel into the bowl ($+1/3$).

\textbf{Pour (4 arms):} The two assistive arms are evaluated on grasping the cup ($+0.5$) and tilting to pour the contents into the target container ($+0.5$). The two primary arms are scored for firmly grasping the central container ($+0.5$) and successfully executing an up-and-down shaking motion to mix the contents ($+0.5$).

\textbf{Spread (4 arms):} All four arms share the identical coordinated criterion: successfully grasping their respective corners of the folded tablecloth ($+0.5$) and synchronously pulling outward to flatten the cloth ($+0.5$).

\textbf{Lift (4 arms):} All four arms share the identical coordinated criterion: achieving a stable grasp on the assigned table leg ($+0.5$) and lifting it to the target height while maintaining planar balance ($+0.5$).

\subsection{Policy Training Hyperparameters}
\label{subsec:hyperparams}
We utilize Diffusion Policy (DP) and 3D Diffusion Policy (DP3) as our imitation learning backbones. The observation spaces and training configurations are detailed below.

\subsubsection{Observation and Action Spaces}
The system receives visual observations from two third-person cameras and proprioceptive data from the robots.

\textbf{Visual Inputs.}
1) DP: The model takes RGB images as input. Images from both cameras are resized to a resolution of $320 \times 240$ pixels.
2) DP3: The model utilizes 3D point cloud representations. Point clouds are captured from two camera views and transformed into a unified world coordinate system using extrinsic calibration matrices. We apply Farthest Point Sampling (FPS) to downsample the point cloud from each camera to $2,048$ points. These are then concatenated to form a dense point cloud of $4,096$ points.
    
\textbf{Proprioception and Action.}
Both architectures condition on a 28-dimensional vector representing the joint positions of the multi-arm system. Similarly, the action space is defined as a 28-dimensional vector corresponding to the target joint positions (and gripper states) for the controlled arms.

\subsubsection{Training Configuration}
The key hyperparameters, including the observation and action horizons, are adjusted based on the complexity and temporal span of the specific task. The ranges for these parameters are listed in Table~\ref{tab:hyperparams}.

\begin{table}[t]
\centering
\caption{Hyperparameters for Policy Training. The horizon parameters are selected within the specified ranges depending on the task difficulty.}
\vspace{0.1cm}
\label{tab:hyperparams}
\renewcommand{\arraystretch}{1.2}
\begin{tabular}{l c}
\toprule
\textbf{Hyperparameter} & \textbf{Value / Range} \\
\midrule
\multicolumn{2}{l}{\textit{Network Inputs}} \\
DP Image Size & $320 \times 240 \times 3$ \\
DP3 Point Cloud Size & $4096 \times 3$ \\
State/Action Dimension & 28 \\
\midrule
\multicolumn{2}{l}{\textit{Training Horizons}} \\
Prediction Horizon ($T_{pred}$) & $16 \sim 32$ \\
Observation Horizon ($T_{obs}$) & $2 \sim 4$ \\
Action Execution Horizon ($T_{act}$) & $8 \sim 16$ \\
\midrule
\multicolumn{2}{l}{\textit{Optimization}} \\
Optimizer & AdamW \\
Learning Rate & $1 \times 10^{-4}$ \\
Batch Size & 64 / 128 \\
Noise Scheduler & DDIM (100 steps) \\
\bottomrule
\end{tabular}
\vspace{-0.6cm}
\end{table}

\section{Prompt Design}
\subsection{System Prompt for LLM Planner}

We design a structured system prompt to guide the planner in decomposing high-level tasks into executable multi-arm stages. The prompt explicitly defines the parallel-sequential execution logic, spatial assignment principles, and available motion primitives. 

Fig.~\ref{fig:system_prompt} presents the full content of the system prompt template. Note that specific object coordinates are dynamically injected at runtime.
Fig.~\ref{fig:plan_output} demonstrates a concrete example of the generated execution plan (JSON format) for a dual-arm manipulation task.

\subsection{VLM Prompting}
\label{subsec:vlm_prompt}

To enable open-vocabulary object manipulation (e.g., ``put items into the gray basket"), we utilize a VLM, specifically Qwen-VL-Max, to ground semantic instructions into 2D image coordinates. Since the VLM output is in 2D pixel space $(u, v)$, we map it to the robot's 3D task space $(X, Y, Z)$ using the aligned depth map from the RealSense camera.

\begin{figure}[h]
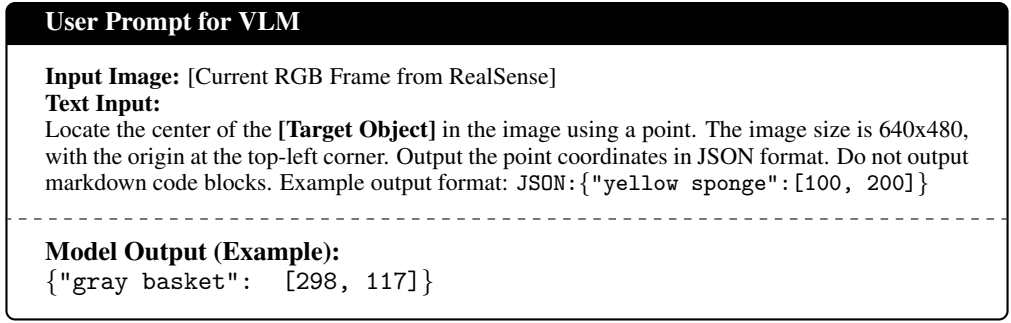

\centering
\begin{tcolorbox}[
    colback=white, 
    colframe=black, 
    boxrule=0.8pt, 
    width=0.95\linewidth,
    title=\textbf{User Prompt for VLM}
]
\small
\textbf{Input Image:} [Current RGB Frame from RealSense] \\
\textbf{Text Input:} \\
Locate the center of the \textbf{[Target Object]} in the image using a point. 
The image size is 640x480, with the origin at the top-left corner. 
Output the point coordinates in JSON format. Do not output markdown code blocks. 
Example output format: \texttt{JSON:\{"yellow sponge":[100, 200]\}}
\tcblower
\textbf{Model Output (Example):} \\
\texttt{\{"gray basket": [298, 117]\}}
\end{tcolorbox}
\vspace{-0.2cm}
\caption{The specific prompt used to query the VLM for object centroids. The model returns 2D pixel coordinates $(u, v)$ corresponding to the target description.}
\label{fig:vlm_prompt}
\end{figure}

We provide the VLM with the current RGB observation and a text prompt requesting the center coordinates of the target object. To ensure the output is machine-readable, we enforce a strict JSON format constraint. 
The prompt template is illustrated in Fig.~\ref{fig:vlm_prompt}.

\subsection{Supervisor Prompting}
\begin{figure}[ht]
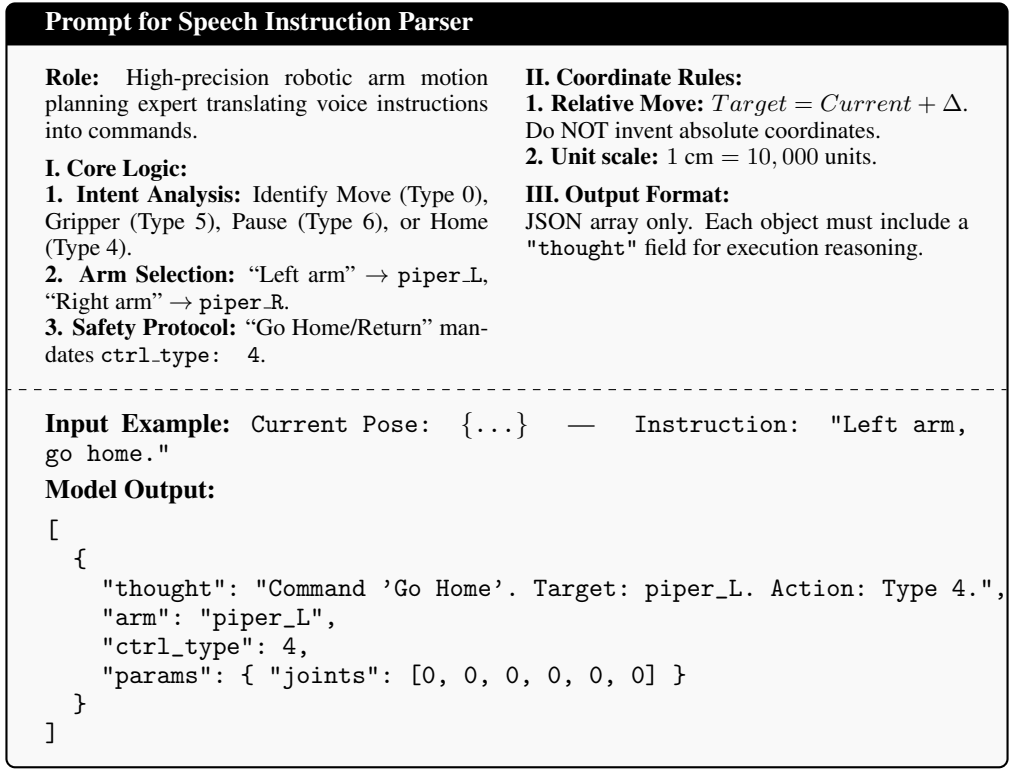
 
\centering
\begin{tcolorbox}[
    colback=gray!5, 
    colframe=black, 
    boxrule=0.8pt, 
    width=0.95\textwidth, 
    title=\textbf{Prompt for Speech Instruction Parser}
]
\footnotesize

\begin{minipage}[t]{0.48\textwidth}
\textbf{Role:} High-precision robotic arm motion planning expert translating voice instructions into commands.

\vspace{0.15cm}
\textbf{I. Core Logic:} \\
\textbf{1. Intent Analysis:} Identify Move (Type 0), Gripper (Type 5), Pause (Type 6), or Home (Type 4). \\
\textbf{2. Arm Selection:} ``Left arm" $\rightarrow$ \texttt{piper\_L}, ``Right arm" $\rightarrow$ \texttt{piper\_R}. \\
\textbf{3. Safety Protocol:} ``Go Home/Return" mandates \texttt{ctrl\_type: 4}.
\end{minipage}
\hfill
\begin{minipage}[t]{0.48\textwidth}
\textbf{II. Coordinate Rules:} \\
\textbf{1. Relative Move:} $Target = Current + \Delta$. Do NOT invent absolute coordinates. \\
\textbf{2. Unit scale:} $1 \text{ cm} = 10,000$ units.

\vspace{0.15cm}
\textbf{III. Output Format:} \\
JSON array only. Each object must include a \texttt{"thought"} field for execution reasoning.
\end{minipage}

\tcblower

\textbf{Input Example:} \texttt{Current Pose: \{...\}} \quad | \quad \texttt{Instruction: "Left arm, go home."}

\vspace{0.1cm}
\textbf{Model Output:}
\begin{verbatim}
[
  { 
    "thought": "Command 'Go Home'. Target: piper_L. Action: Type 4.",
    "arm": "piper_L", 
    "ctrl_type": 4, 
    "params": { "joints": [0, 0, 0, 0, 0, 0] } 
  }
]
\end{verbatim}

\end{tcolorbox}
\vspace{-0.3cm}
\caption{The system prompt for the Voice Command Supervisor. The example demonstrates the handling of a "Go Home" command, which triggers a safety reset primitive (Type 4) to bring the robotic arm back to its initial position.}
\label{fig:voice_prompt}
\vspace{-0.2cm}
\end{figure}

Unlike the high-level task planner, the voice supervisor requires precise, immediate, and deterministic responses. We design a specialized prompt that strictly maps natural language intent to control primitives.
As shown in Fig.~\ref{fig:voice_prompt}, the prompt enforces a ``Chain of Thought" process to ensure that critical commands like \textbf{``Go Home"} are correctly identified and mapped to the safety reset primitive (Type 4), overriding any ongoing trajectory.

\begin{figure}[ht] 
\centering
\begin{tcolorbox}[
    colback=gray!5, 
    colframe=black, 
    boxrule=0.8pt, 
    width=0.95\textwidth, 
    title=\textbf{System Prompt Template for Planner}
]
\footnotesize 

\begin{minipage}[t]{0.48\textwidth}
\textbf{I. Role \& Goal:} You are a multi-robotic arm collaboration task planning assistant. Your goal is to decompose the task into executable \textbf{Stages} for two arms: \texttt{"piper\_L"} and \texttt{"piper\_R"}.

\vspace{0.15cm}
\textbf{II. Task:} I want to clear the table.

\vspace{0.15cm}
\textbf{III. Execution Logic:} \\
\textbf{1. Sequential Stages:} Complete all actions in current stage before the next. \\
\textbf{2. Parallel Actions:} Actions within the same stage execute simultaneously.

\vspace{0.15cm}
\textbf{IV. Task Assignment Principles:} \\
\texttt{piper\_L}: Targets with \textbf{positive X} ($X > 0$). \\
\texttt{piper\_R}: Targets with \textbf{negative X} ($X < 0$).
\end{minipage}
\hfill
\begin{minipage}[t]{0.48\textwidth}
\textbf{V. Motion Primitives (ctrl\_type):} \\
\texttt{2 (Pick)}: Move to target and grasp. Params: $\{X, Y, Z\}$. \\
\texttt{3 (Place)}: Move to target and release. Params: $\{X, Y, Z\}$. \\
\texttt{4 (Home)}: Return to origin. Params: $\{joints: [0, \dots, 0]\}$.

\vspace{0.15cm}
\textbf{VI. Collision Avoidance Constraints:} \\
If sharing a single place point, the first arriving arm must return Home immediately to clear the workspace. If distinct points exist, arms proceed in parallel.

\vspace{0.15cm}
\textbf{VII. Strict Output Format (JSON):} \\
\texttt{[\{ "arm": "piper\_L", "ctrl\_type": 2, "params": \{...\} \},} \\
\texttt{ \{ "arm": "piper\_R", "ctrl\_type": 3, "params": \{...\} \}]}
\end{minipage}

\end{tcolorbox}
\vspace{-0.3cm}
\caption{The structured system prompt used for the LLM planner. It enforces spatial constraints and collision avoidance rules.}
\label{fig:system_prompt}
\end{figure}
\vspace{-0.5cm}

\begin{figure}[h]
\centering
\begin{tcolorbox}[
    colback=gray!5, 
    colframe=black, 
    boxrule=0.8pt, 
    width=0.95\textwidth, 
    title=\textbf{Example of Plan Output}
]
\footnotesize 
\begin{verbatim}
plan = [
  # Stage 1: Parallel Picking (Both arms move simultaneously)
  [ {'arm': 'piper_L', 'executor': 'agent', 'ctrl_type': 2, 
     'params': {'X': 0.135, 'Y': -0.069, 'Z': 0.501}},
    {'arm': 'piper_R', 'executor': 'agent', 'ctrl_type': 2, 
     'params': {'X': -0.151, 'Y': 0.087, 'Z': 0.471}} ],

  # Stage 2: Left Arm Place (Sequential execution to avoid collision)
  [ {'arm': 'piper_L', 'executor': 'agent', 'ctrl_type': 3, 
     'params': {'X': 0.017, 'Y': -0.123, 'Z': 0.541}} ],

  # Stage 3: Left Arm Returns Home (Clears workspace)
  [ {'arm': 'piper_L', 'executor': 'agent', 'ctrl_type': 4, 
     'params': {'joints': [0, 0, 0, 0, 0, 0]}} ],

  # Stage 4: Right Arm Place (Starts only after Stage 3 is complete)
  [ {'arm': 'piper_R', 'executor': 'agent', 'ctrl_type': 3, 
     'params': {'X': 0.017, 'Y': -0.123, 'Z': 0.541}} ],

  # Stage 5: Right Arm Returns Home
  [ {'arm': 'piper_R', 'executor': 'agent', 'ctrl_type': 4, 
     'params': {'joints': [0, 0, 0, 0, 0, 0]}} ]
]
\end{verbatim}
\end{tcolorbox}
\vspace{-0.3cm}
\caption{A concrete example of the output generated by the LLM planner. The plan consists of sequential stages, where actions within each stage are executed in parallel.}
\label{fig:plan_output}
\vspace{-0.2cm}
\end{figure}

\section{Additional Experiments} 
\subsection{Trajectory Quality Analysis}
\label{sec:trajectory_quality_analysis}

High-quality demonstrations are typically characterized by low variance and smooth motion trajectories, which are beneficial for downstream imitation learning \cite{argall2009survey, ravichandar2020recent}. These qualities reduce the multimodality and noise that learned policies must account for, thereby facilitating more stable and accurate policy convergence \cite{mandlekar2021matters}.

\subsubsection{Spatial Consistency Across Tasks}
\begin{figure}[htbp]
    \centering

    \begin{subfigure}[b]{0.32\textwidth}
        \centering
        \includegraphics[width=\textwidth]{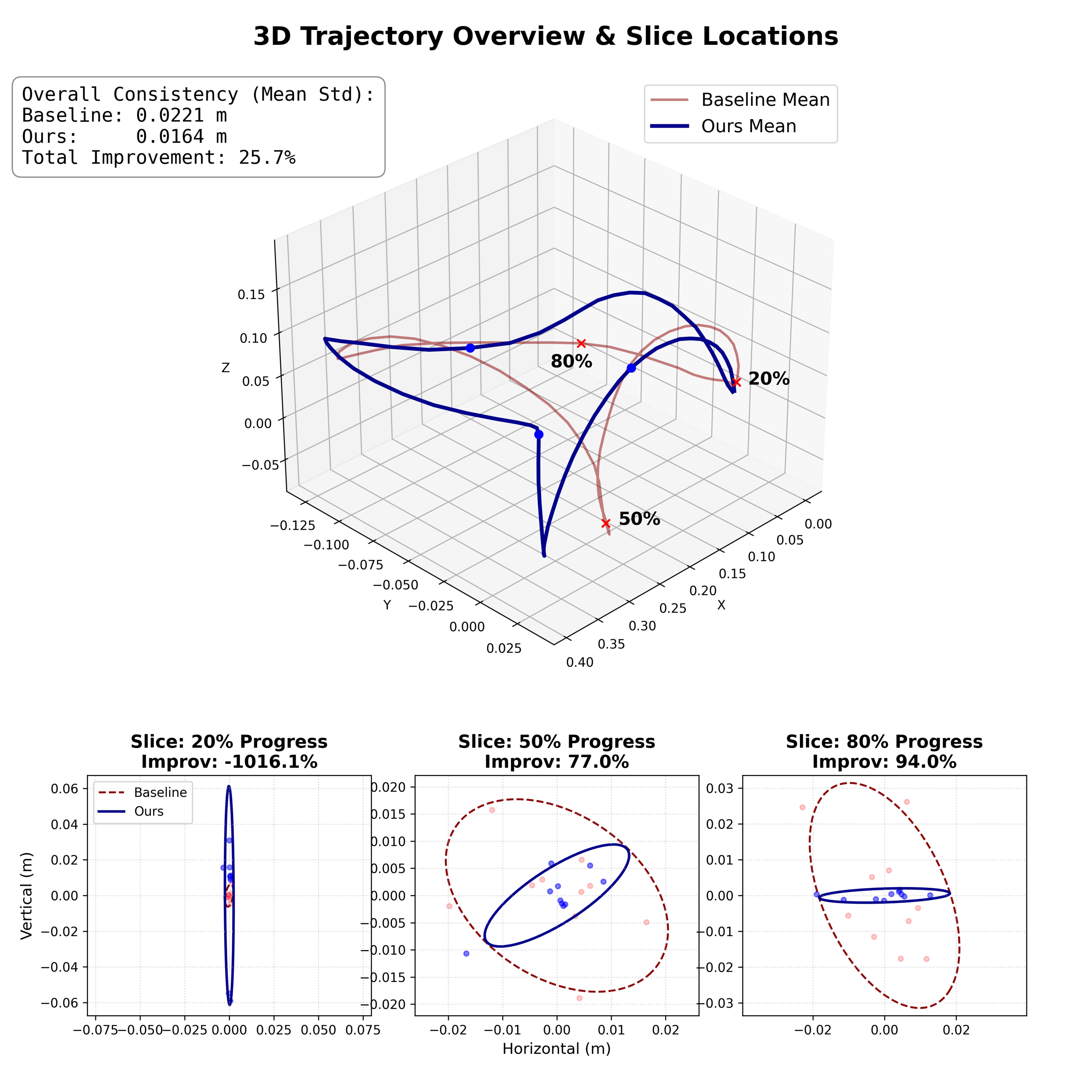}
        \vspace{-0.5cm}
        \caption{Clear}
        \label{fig:traj_clear}
    \end{subfigure}\hfill
    \begin{subfigure}[b]{0.32\textwidth}
        \centering
        \includegraphics[width=\textwidth]{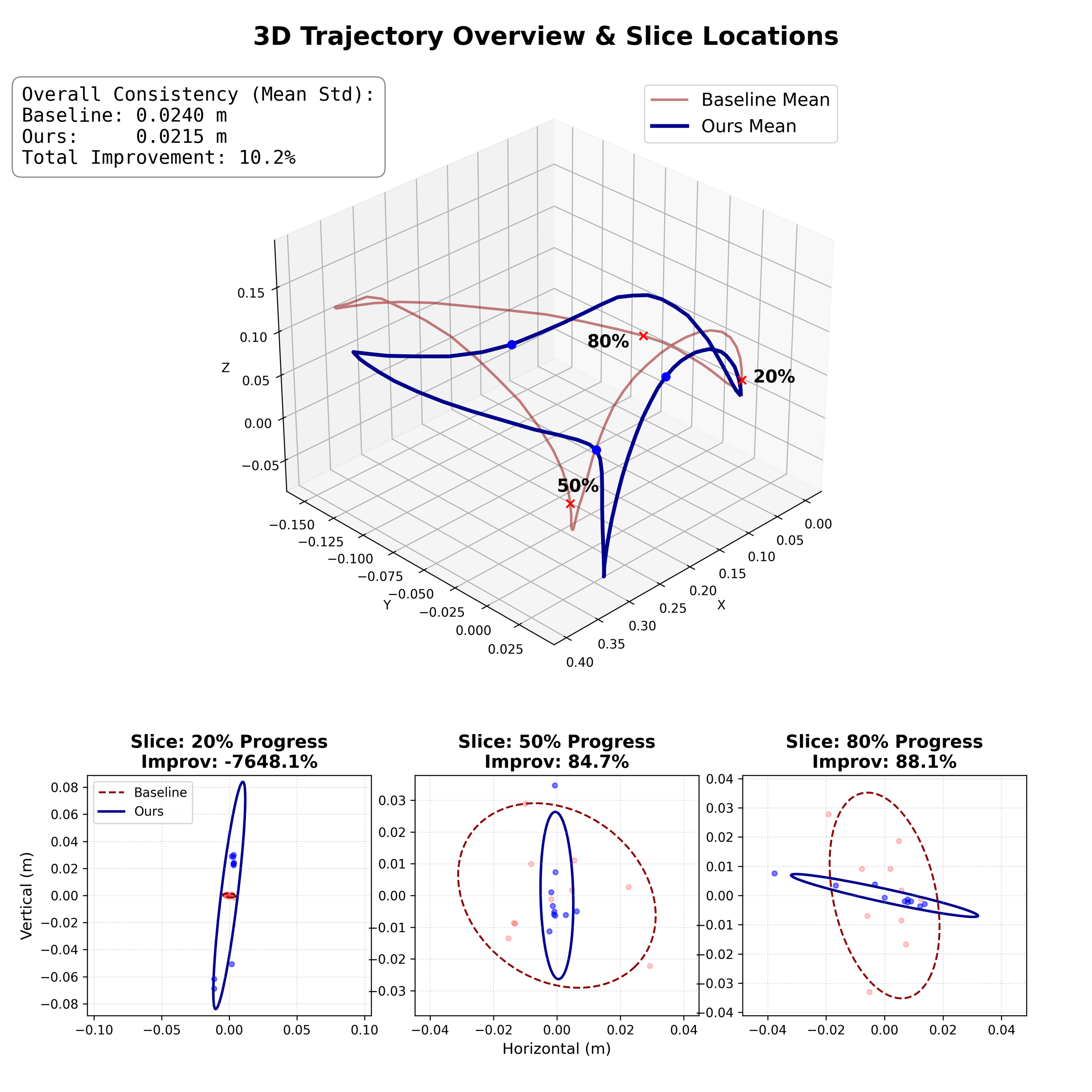}
        \vspace{-0.5cm}
        \caption{Pack}
        \label{fig:traj_pack}
    \end{subfigure}\hfill
    \begin{subfigure}[b]{0.32\textwidth}
        \centering
        \includegraphics[width=\textwidth]{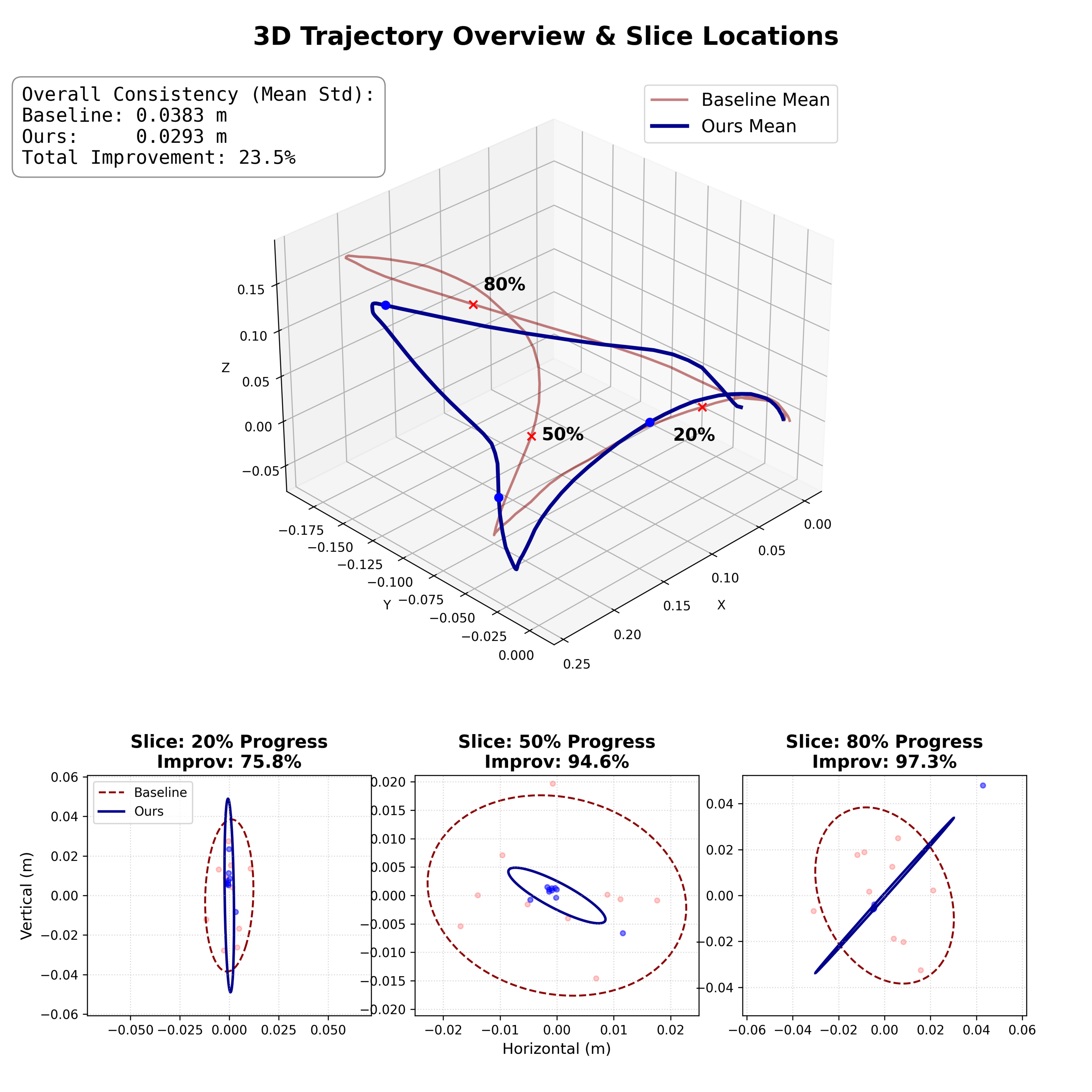}
        \vspace{-0.5cm}
        \caption{Draw}
        \label{fig:traj_draw}
    \end{subfigure}
    
    \vspace{0.2cm} 

    \begin{subfigure}[b]{0.32\textwidth}
        \centering
        \includegraphics[width=\textwidth]{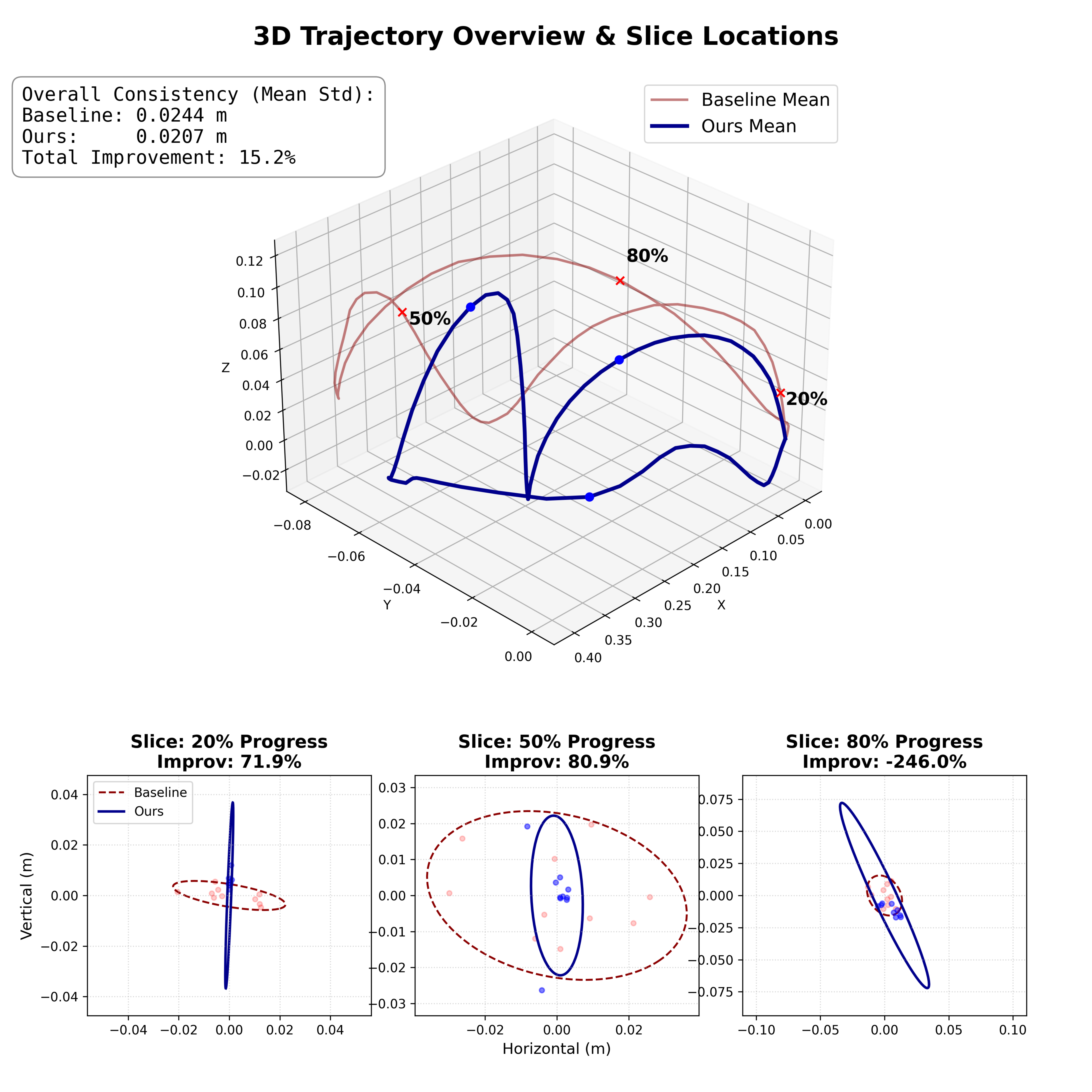}
        \vspace{-0.5cm}
        \caption{Clean}
        \label{fig:traj_clean}
    \end{subfigure}\hfill
    \begin{subfigure}[b]{0.32\textwidth}
        \centering
        \includegraphics[width=\textwidth]{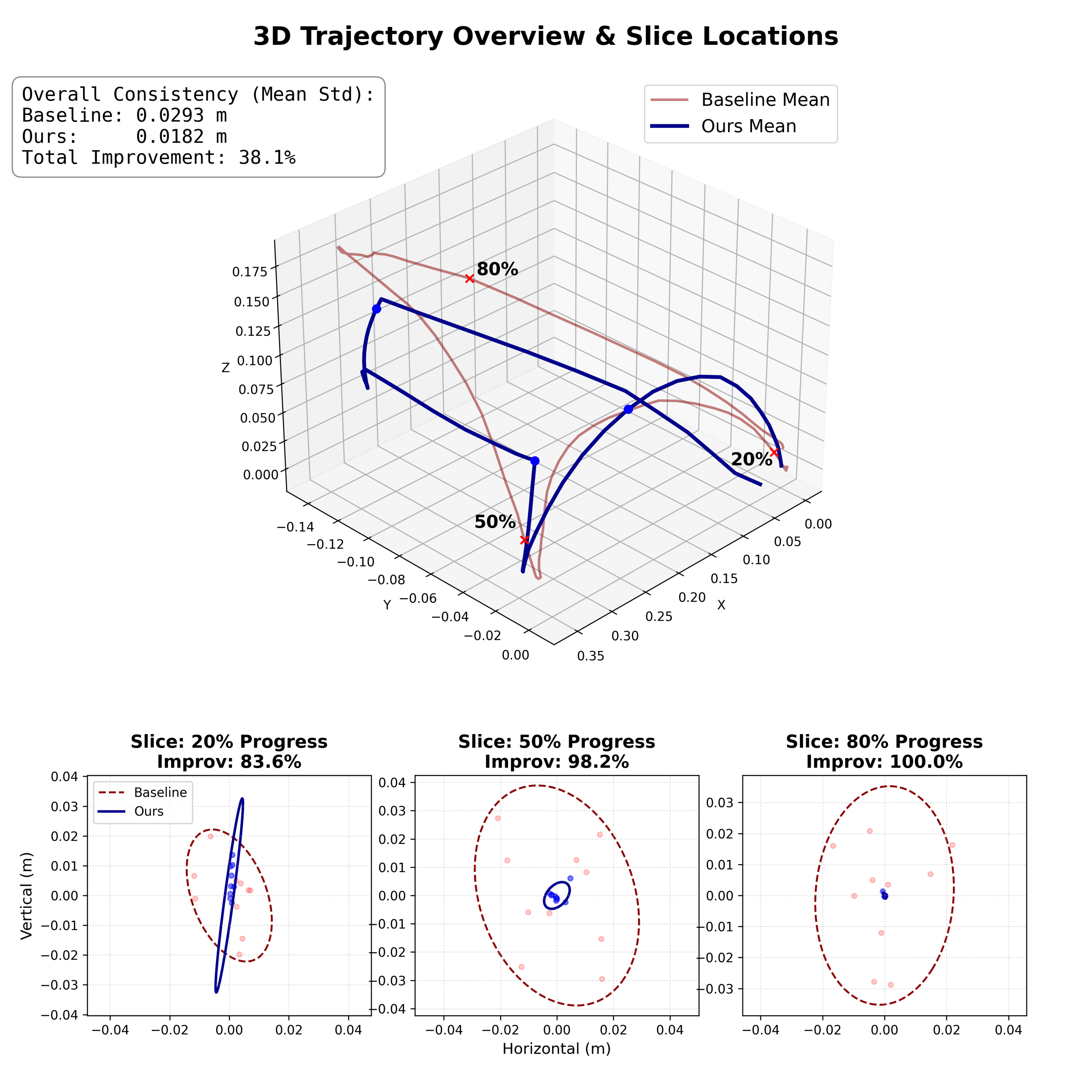}
        \vspace{-0.5cm}
        \caption{Pour}
        \label{fig:traj_pour}
    \end{subfigure}\hfill
    \begin{subfigure}[b]{0.32\textwidth}
        \centering
        \includegraphics[width=\textwidth]{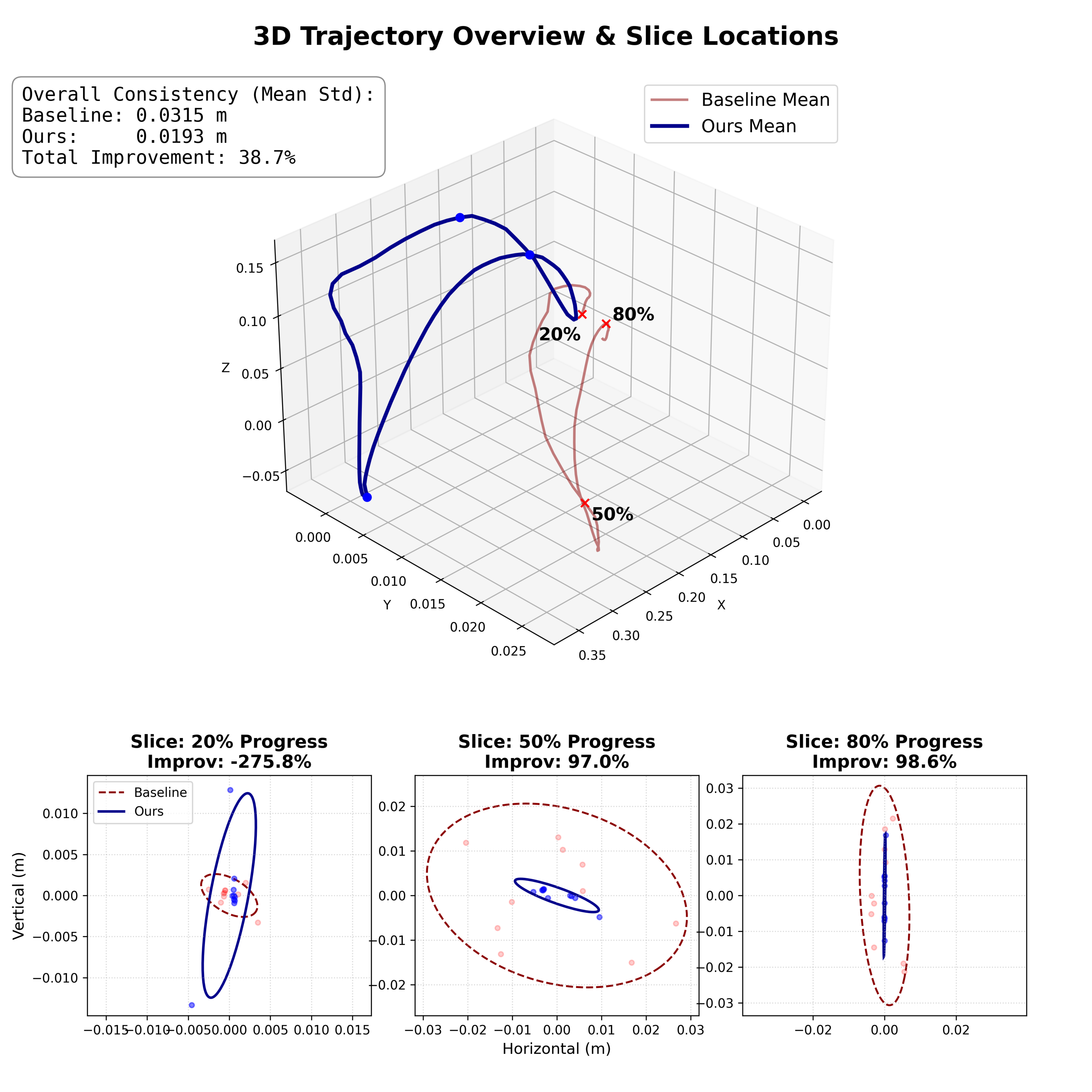}
        \vspace{-0.5cm}
        \caption{Spread}
        \label{fig:traj_spread}
    \end{subfigure}
    
    \vspace{0.2cm} 

    \begin{subfigure}[b]{0.32\textwidth}
        \centering
        \includegraphics[width=\textwidth]{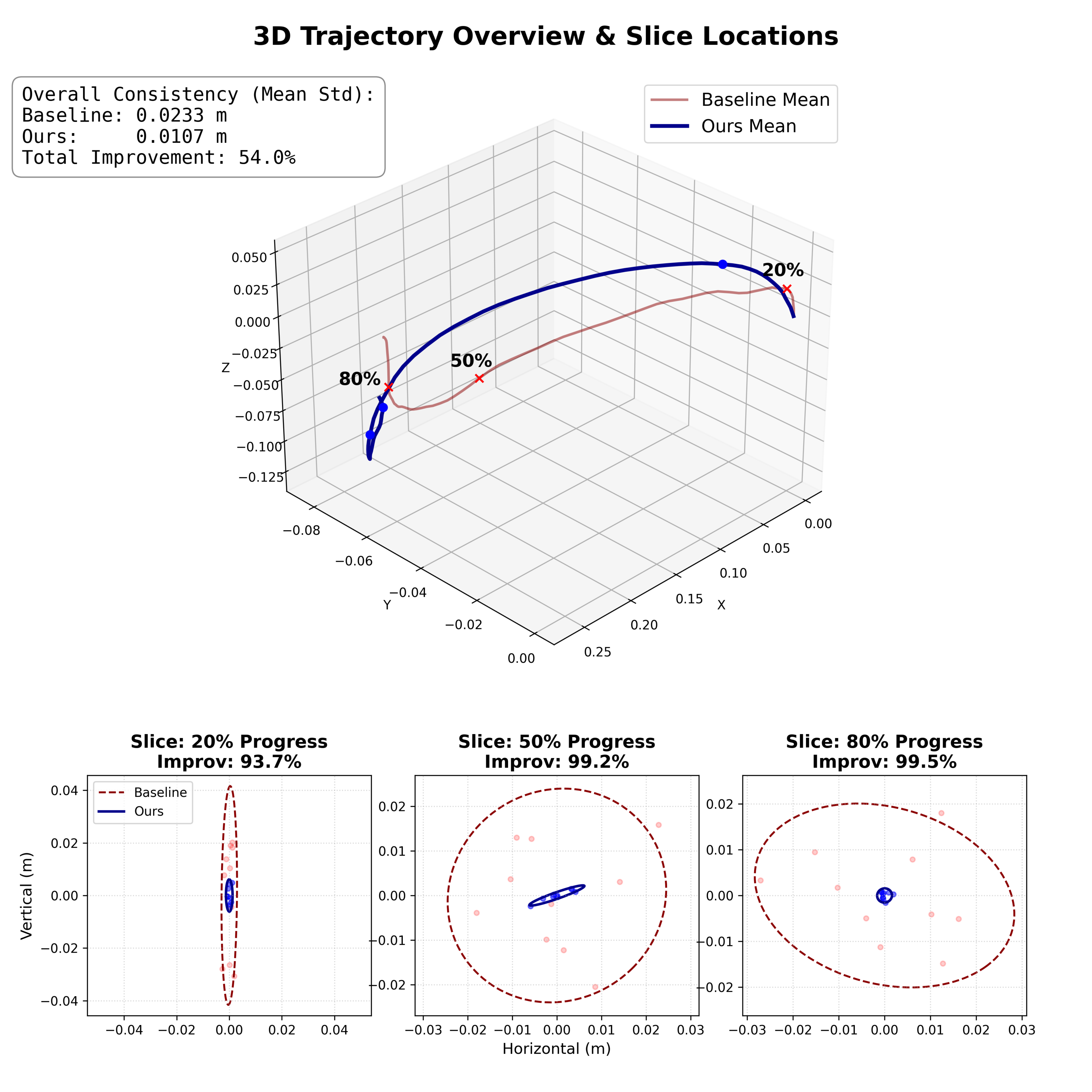}
        \vspace{-0.5cm}
        \caption{Lift}
        \label{fig:traj_lift}
    \end{subfigure}
    
    \vspace{-0.1cm}
    \caption{Trajectory Analysis across all seven tasks. Top: 3D visualization of the mean trajectories. Bottom: Cross-sectional scatter plots at 20\%, 50\%, and 80\% of the trajectory progress. The blue ellipses (HATS) are significantly tighter than the red ellipses (Dual-Human baseline), indicating substantially higher spatial precision and consistency.}
    \label{fig:all_trajectories} 
    \vspace{-0.4cm}
\end{figure}

To evaluate data consistency, we perform a detailed comparative analysis using the 3D end-effector trajectories of a selected single assistive robotic arm across 20 repeated execution trials for each task. As visualized in Fig.~\ref{fig:all_trajectories}, HATS (blue trajectories) follows the planned motion primitives closely, whereas the unassisted Dual-Human baseline (red trajectories) exhibits higher spatial variability due to natural human motor noise and coordination drift.

Quantitatively, HATS consistently reduces the Mean Spatial Standard Deviation across all seven tasks. The most significant gains are observed in structurally constrained and strongly coupled tasks, including \textit{Lift} (54.0\% improvement, Fig.~\ref{fig:all_trajectories}g), \textit{Spread} (38.7\%, Fig.~\ref{fig:all_trajectories}f), and \textit{Pour} (38.1\%, Fig.~\ref{fig:all_trajectories}e). For tasks requiring complex but collision-free geometric routing, such as \textit{Clear} (Fig.~\ref{fig:all_trajectories}a) and \textit{Draw} (Fig.~\ref{fig:all_trajectories}c), HATS maintains strong improvements of 25.7\% and 23.5\%, respectively. Furthermore, even in high-frequency, contact-rich manipulation scenarios like \textit{Clean} (Fig.~\ref{fig:all_trajectories}d) and \textit{Pack} (Fig.~\ref{fig:all_trajectories}b), where baseline variance is naturally elevated due to physical interactions, HATS successfully enforces a tighter distribution, yielding improvements of 15.2\% and 10.2\%.

\subsubsection{Cross-Sectional Dispersion Analysis}
To further quantify this spatial dispersion, we extract cross-sectional scatter plots at 20\%, 50\%, and 80\% of the task progress (as shown in the bottom panels of Fig.~\ref{fig:all_trajectories}). 
Across the critical mid-to-late execution phases (50\% and 80\% progress), the 95\% confidence ellipses of HATS are markedly smaller and more concentrated than those of the baseline, frequently demonstrating over 80\% local improvement in spatial precision. 
This structural stability suggests that the agent effectively mitigates the execution jitter inherent in human teleoperation. By minimizing the spatial distribution shift in the demonstration dataset, HATS therefore provides a more spatially consistent demonstration distribution for downstream policy training.

\subsubsection{Kinematic Smoothness}
In addition to spatial consistency, we assess motion quality using the Mean Jerk metric \cite{flash1985coordination}, where lower values indicate more fluid movements. Despite the discrete nature of the primitive-based planning for assistive arms, both the Dual-Human baseline and HATS exhibit similarly low mean jerk values (approximately $0.001$ on average across tasks). This indicates that the agent's motions remain as fluid and naturalistic as human experts. These results indicate that HATS can generate spatially consistent and smooth demonstrations suitable for downstream multi-arm policy learning.

\subsection{Ablation Study on LLM Planner}

To validate the necessity of our LLM-driven planning module, we compare HATS against a traditional \textbf{Rule-based Scheduler}. The baseline relies on manual task assignment and fixed execution sequences without replanning capabilities. 

\textbf{Adaptability to scene variations.} We dynamically alter the \textit{Clear} task (initially cached for $N=2$ objects). As shown in Table~\ref{tab:planner_ablation}, the rule-based system fails in both edge cases. If an object is added ($N=3$), it rigidly executes only the two programmed stages, leaving the third object untouched. If an object is removed ($N=1$), it mindlessly proceeds to the pre-programmed second stage, performing empty placements without an actual target. Conversely, HATS leverages its perception-aware validator to detect these mismatches, automatically triggering a \textit{Replan} to adjust stage allocations and achieving 100\% completion.

\textbf{Deployment scalability.} Expanding a rule-based system to novel tasks is notoriously tedious, requiring researchers to manually script role assignments and rigid collision locks. Crucially, as the task horizon lengthens (i.e., requiring more execution steps), the complexity of hardcoding these temporal dependencies scales exponentially, demanding prohibitive human effort. HATS eliminates this bottleneck. By acting as a zero-shot compiler, it simply takes a high-level natural language prompt and autonomously generates collision-free primitive sequences, making the deployment of complex, long-horizon tasks effortless and highly scalable.

\subsection{Failure Analysis}

\begin{wrapfigure}{r}{0.45\textwidth}
  \vspace{-0.6cm} 
  \centering
  \includegraphics[width=\linewidth]{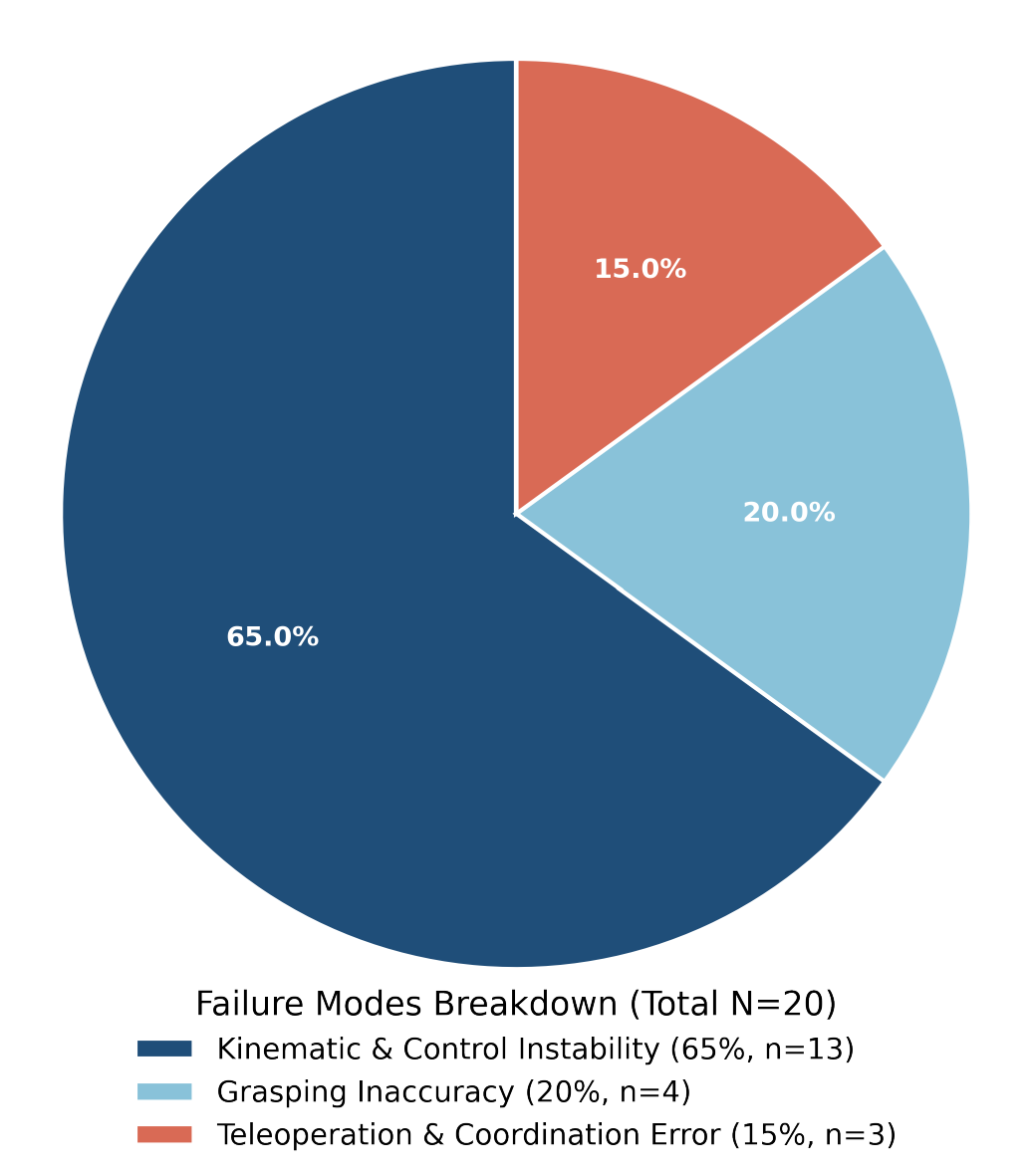}
  \caption{Distribution of failure modes analyzed from 20 execution failures. Hardware kinematic instability is the dominant factor.}
  \label{fig:failure_pie}
  \vspace{-0.3cm} 
\end{wrapfigure}

To identify the limitations of our system, we conducted an analysis of 20 failure cases observed during the experimental evaluation. As illustrated in Fig.~\ref{fig:failure_pie}, the failures are categorized into three primary sources:

\begin{table}[t]
  \centering
  \small
  \caption{Ablation on Planning Module.}
  \label{tab:planner_ablation}
  \renewcommand{\arraystretch}{1.1}
  \setlength{\tabcolsep}{4pt} 
  \begin{tabular}{lccc}
    \toprule
    \multirow{2}{*}{\textbf{Method}} & \multicolumn{2}{c}{\textbf{Scene Variation (Cached $N=2$)}} & \multirow{2}{*}{\textbf{Deployment Effort}} \\
    \cmidrule(lr){2-3}
    & \textbf{Added Object ($N=3$)} & \textbf{Removed Object ($N=1$)} & \\
    \midrule
    Rule-based & Fails  & Fails  & High  \\
    \textbf{LLM (Ours)} & \textbf{Succeeds} & \textbf{Succeeds} & \textbf{Low} \\
    \bottomrule
  \end{tabular}
  \vspace{-0.5cm}
\end{table}

\textbf{Kinematic and control instability (65\%).}
The majority of failures stem from the low-level control limitations of the hardware. The Piper manipulators rely on an internal Inverse Kinematics (IK) solver to convert Cartesian target poses (from \texttt{EndPoseCtrl}) into joint angles. We observed that this internal solver occasionally converges to unreasonable joint configurations or hits singularities, causing the arm to stutter or fail to reach the target pose. Crucially, this behavior can be nondeterministic; the same command may succeed in one trial and fail in another due to the arm's initial configuration. To reduce singularity-induced failures, we incorporate pre-defined grasping priors for objects with known geometries, ensuring the target end-effector orientation is within a feasible kinematic workspace. 

Importantly, our framework is hardware-agnostic. The observed kinematic instabilities are intrinsic to the low-cost manipulators used in this study. We anticipate that deploying our system on industrial-grade robots with more robust IK solvers (e.g., Franka) could potentially mitigate these failures.

\textbf{Grasping inaccuracy (20\%).}
This category involves failures where the gripper fails to secure the object despite reaching the target vicinity. The robotic gripper has a limited stroke of approximately 70 mm. When the grasp pose generated by AnyGrasp has a positional error—even a minor deviation—it can result in a collision with the object or a grasp attempt on a section wider than the gripper limit, particularly for small or irregularly shaped objects.

\textbf{Teleoperation and coordination error (15\%).} 
Human-related errors account for the remaining failures. These occur primarily in: (1) Synchronization Latency: In tightly coupled tasks (e.g., \textit{Lift} and \textit{Spread}), the human operator may react slower than the autonomous agent, disrupting the required dual-arm coordination. (2) Operation Difficulty: Fine manipulation tasks, such as opening a drawer or closing a lid (in the \textit{Pack} task), impose a high cognitive load, leading to occasional manual control errors.


\end{document}